\documentclass{article}
\usepackage[preprint]{neurips_2026}
\usepackage[utf8]{inputenc}
\usepackage[T1]{fontenc}
\usepackage{microtype}

\usepackage{amsmath,amsfonts,bm}

\def\eqref#1{equation~\ref{#1}}

\def\1{\bm{1}}

\DeclareMathAlphabet{\mathsfit}{\encodingdefault}{\sfdefault}{m}{sl}
\SetMathAlphabet{\mathsfit}{bold}{\encodingdefault}{\sfdefault}{bx}{n}

\usepackage{hyperref}
\usepackage{url}
\usepackage{graphicx}
\usepackage{wrapfig}
\usepackage{booktabs}
\usepackage{placeins}

\title{Learning Polarization Image Restoration with General Restoration Priors}

\author{%
  Chenggong Li \\
  School of Automation \\
  Central South University \\
  \texttt{244603040@csu.edu.cn}
  \And
  Jinhao Liu \\
  School of Automation \\
  Central South University
  \AND
  Caiyun Wu \\
  School of Automation \\
  Central South University
  \And
  Yidong Luo \\
  School of Engineering \\
  Westlake University
  \AND
  Junchao Zhang \\
  School of Automation \\
  Central South University\\
  \texttt{junchaozhang@csu.edu.cn}
  \And
  Degui Yang \\
  School of Automation \\
  Central South University\\
  \texttt{degui.yang@csu.edu.cn}
}

\begin{document}

\maketitle

\begin{abstract}
Polarization imaging captures distinctive surface and
geometric cues that benefit a wide range of vision tasks.
However, real-world polarization acquisition is often
affected by multiple coupled degradations, making image
restoration essential for practical polarization vision.
Existing methods are largely tailored to specific
degradations and remain constrained by the limited
scale and quality of polarization data. To address
these limitations, we develop an all-in-one polarization restoration
framework for diverse and composite degradations. We first study the impact of different
polarization representations on restoration performance and identify the normalized
Stokes representation as an effective choice for separating
intensity and polarization information. Accordingly, we devise a
dual-branch architecture that separates intensity and
polarization modeling.
To overcome the limitations of polarization-specific training,
the intensity branch leverages pretrained general restoration
priors and a mixture-of-experts extension for composite degradations,
while its restoration knowledge is adaptively distilled into
the symmetric polarization branch via a cross-domain feature transform.
In addition, we establish a composite-degradation polarization
benchmark to support all-in-one restoration research.
Extensive experiments on public datasets and our
proposed benchmark demonstrate the effectiveness
of the proposed method.
\end{abstract}

\section{Introdution}
Polarization imaging reveals unique physical properties of a scene,
such as reflectance, surface roughness, and geometry.
These polarimetric cues benefit various downstream tasks,
including object detection (\cite{luo2025cpifuse}), semantic segmentation (\cite{liu2025sharecmp}), reflection removal (\cite{yao2025polarfree,lyu2022physics,lei2020polarized,wang2026polarization}),
and surface normal estimation (\cite{deschaintre2021deep,lyu2023shape,lei2022shape}). However, the key to fully leveraging these benefits lies in the accurate acquisition
of polarization properties. Polarization imaging mainly follows two paradigms,
division-of-time (DoT) and division-of-focal-plane (DoFP), both of which acquire
multi-angle intensity measurements to derive the degree of linear polarization (DoLP) and angle of polarization (AoP) (\cite{rebhan2019principle}). This imaging mechanism makes
polarization images more vulnerable to real-world degradations, including
blur, low illumination, and noise, which limits their practical applications.
Therefore, polarization image restoration is crucial for reliable deployment of polarization imaging.  

\begin{figure}[!t]
	\begin{center}
		\includegraphics[width=\linewidth]{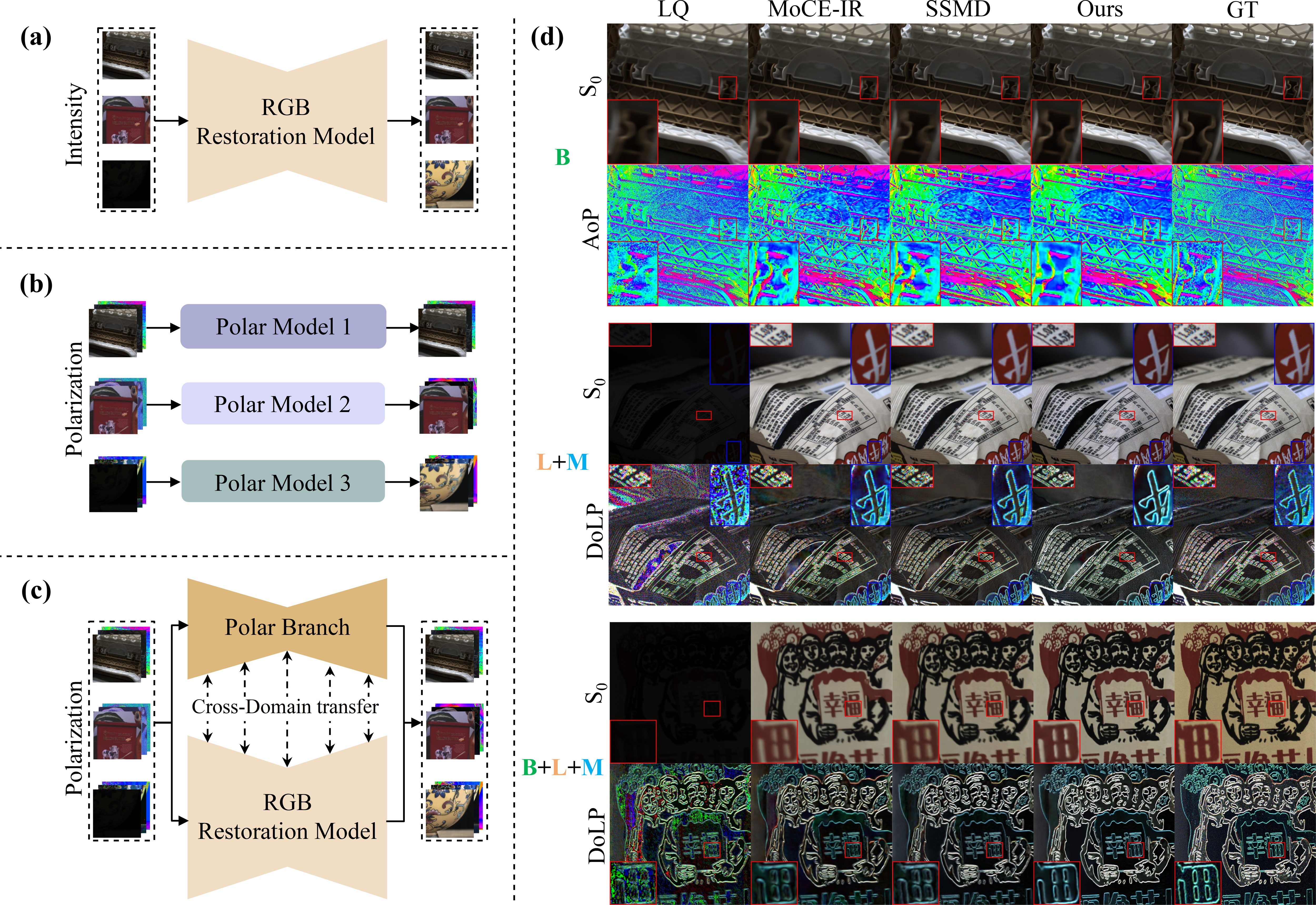}
	\end{center}
	\vspace{-0.8em}
	\caption{Polarization restoration approach. (a) Angle-wise general
	restoration overlooks polarization correlations.
	(b) Degradation-specific experts (``Polar Model'') handle only individual degradations
	and remain constrained by polarization data distribution.
	(c) Our framework combines general restoration priors
	with polarization-specific modeling for
	restoration across diverse degradations.
	(d) Visual comparisons on blur (B), low light (L), mosaic (M),
	and their combinations: low quality image (LQ), MoCE-IR (\cite{zamfir2025complexity}), SSMD (\cite{zhou2026architectural}), Ours, and ground truth (GT) from
	left to right.}
	\label{fig:1}
	\vspace{-1.5 em}
\end{figure}

The performance of image restoration models is largely determined by
the quality of the training data. Existing polarization image restoration
methods typically train models using paired polarization image datasets,
either synthesized (\cite{zhou2025learning}) or captured from real
scenes
(\cite{zhou2023polarization,xu2022colorpolarnet,rahman2025polarization,li2026polarvsr}).
Although these datasets enable promising restoration performance,
they still lag far behind general image restoration datasets.
Taking image deblurring as an example, conventional image deblurring
methods usually construct blurred-sharp image pairs through
frame averaging from high-frame-rate videos, such as the
GoPro (\cite{nah2017deep}) and REDS (\cite{nah2019ntire}) datasets.
However, due to the limitations of polarization cameras,
existing polarization deblurring methods mainly rely on synthetic
blur kernels to generate blurred-sharp pairs (\cite{zhou2025learning}).
Moreover, the complex acquisition process of polarization data
further restricts the scale and diversity of polarization
datasets. Meanwhile, models carefully designed and trained on
general restoration datasets have demonstrated strong performance and
generalization, naturally raising a question: \textit{Can
general restoration priors be effectively transferred to
polarization image restoration to overcome exisiting domain-specific limitations?}

Beyond data, existing methods also face limitations in
model design.
Although current approaches optimize restoration
networks towards the final polarization image quality,
the representations learned by these models are largely
chosen empirically, spanning the intensity domain (\cite{li2025demosaicking,luo2024learning}),
polarization domain (\cite{zhou2023polarization}), or combinations of both (\cite{xu2022colorpolarnet,zhou2026architectural}).
The impact of different polarization representations
on restoration performance has received limited attention. This
leads to another fundamental question: \textit{Which representation is better suited
for polarization image restoration?} Furthermore, existing studies mainly focus
on developing task-specific models with dedicated
parameters for individual polarization degradations.
In contrast, an all-in-one (AIO) polarization image
restoration framework capable of handling diverse
degradations remains largely unexplored.

Based on these observations, we propose to leverage general
restoration priors for addressing polarization degradation.
Specifically, we first conduct a systematic investigation into
the impact of different polarization representations on restoration
performance, and reveal that normalized Stokes-based
representations are more suitable for diverse restoration tasks due to its
intensity-polarization decoupling property.
Based on this finding, we develop a dual-branch restoration framework
that separately models intensity and polarization information.
For the intensity branch, we adapt a pretrained general restoration
model (\cite{conde2024instructir}) and extend it into a mixture-of-experts (MoE) architecture (\cite{puigcerver2024sparse}) to better
handle complex degradations, while introducing a symmetric polarization
branch to recover polarization-specific information.
To bridge general restoration priors and
polarization-specific representations, we introduce
a bidirectional interaction mechanism. A adaptive cross-domain feature transform
selectively
distills transferable intensity priors to guide polarization restoration,
while complementary polarization cues
are fed back via zero-convolution modulation to preserve the
pretrained intensity restoration capability.
The proposed dual-branch architecture preserves domain-specific modeling to retain general restoration priors,
while enabling effective information exchange between intensity and polarization representations.
Finally, we introduce PolarCDD, a new polarization restoration
dataset covering 11 degradation settings composed of blur,
low light, noise, mosaic artifacts, and their combinations,
to facilitate research on AIO and composite
polarization restoration.

In summary, our contributions can be summarized as follows:
    (1) We investigate how different representation
	choices affect polarization restoration and find
	that the normalized Stokes representation provides the best
	trade-off for polarization fidelity.
    (2) We develop an AIO polarization restoration
	model that equips the intensity branch with
	pretrained restoration priors and an MoE extension,
	while leveraging an adaptive modulation mechanism
	to transfer its restoration capability to the polarization domain.
    (3) We establish PolarCDD, a dataset with diverse composite polarization degradations, to facilitate future research in polarization image restoration.

\section{Related Work}
\label{gen_inst}

\textbf{Polarization Image Restoration.} Most polarization image restoration methods
are task-specific for individual degradations, and typically exploit either
intensity information, polarization cues, or their complementary
relationship. For example, intensity-guided restoration has been
explored for polarization deblurring and enhancement, while dual-domain
or parallel designs further model intensity and polarization jointly
to improve fidelity
(\cite{zhou2025learning,xu2022colorpolarnet,zhou2023polarization}).
For denoising, prior work has explored residual dense feature aggregation
for polarization parameter recovery and hybrid-attention Transformers
with physics-based noise modeling for
low-light polarization images (\cite{li2020learning,li2023polarized}). For demosaicking,
prior work has investigated staged reconstruction, adversarial
learning, diffusion models, and joint optimization with downstream tasks
(\cite{nguyen2022two,guo2024attention,li2025demosaicking,luo2026polarapp}). More recently, Zhou et al.
investigate general network architectures for polarization
image restoration and propose a single-stage dual-domain
design applicable to different restoration tasks
(\cite{zhou2026architectural}).

Despite this progress, existing methods remain largely
constrained by limited polarization data and task-specific architectural
design. And the choice of polarization
representation—an essential factor that determines
how intensity and polarization information are encoded
and restored—has received little systematic investigation.
In addition, AIO restoration for diverse polarization
degradations remains largely unexplored.

\textbf{General Image Restoration.} General image restoration
has progressed from task-specific backbones toward unified
models capable of handling diverse degradations.
Representative architectures such as Restormer and NAFNet
establish strong generic restoration backbones through
efficient attention and simplified network design
(\cite{zamir2022restormer,chen2022simple}). Building
upon these backbones, recent AIO methods
introduce degradation-aware conditioning or
adaptive routing to accommodate multiple restoration tasks.
PromptIR and InstructIR employ learned prompts or textual
instructions to condition restoration behavior
(\cite{potlapalli2023promptir,conde2024instructir}), while
OneRestore and AdaIR model degradation characteristics through
multimodal or frequency-aware representations
(\cite{guo2024onerestore,cui2025adair}). MoCE-IR further
introduces mixture-of-experts routing to dynamically
allocate restoration capacity across degradations
(\cite{zamfir2025complexity}). Beyond discriminative models,
recent work has explored large-scale generative and
foundation priors for image restoration, demonstrating
the benefit of knowledge learned from broader restoration
data (\cite{fei2026lucidflux,chen2026foundir}). However,
these advances are primarily developed for conventional RGB
images. This motivates us to investigate whether general
restoration priors can be effectively transferred to
polarization restoration, where data are substantially
more limited and both intensity fidelity and polarization
accuracy must be preserved.

\section{METHODOLOGY}
\label{headings}
\subsection{Preliminaries}

Polarization imaging captures intensity images at
multiple polarization angles and subsequently derives
the polarization properties of the scene. Specifically,
the polarization image $x_{\theta}$ at a given angle $\theta$ can be
represented as:
\begin{equation}
    \begin{split}
{x_\theta } = I\{ 1 + p\cos 2(\theta  - \phi )\} ,
\end{split}
    \label{eq:1}
\end{equation}
where $I$ is the unpolarized light intensity, $p$ and $\phi$ denote the degree and angle of
polarization (DoLP and AoP), respectively. $p$ and $\phi$
are governed by the material properties, surface geometry, and viewing direction
, and thus encode rich polarization cues of the scene. We represent the
captured intensity measurements at all polarization
angles as ${X} = [ x_{{0^ \circ }},x_{{{45}^ \circ }},x_{{{90}^ \circ }},x_{{{135}^ \circ }}]$.
$X$ can be further transformed into a more suitable representation for polarization imaging, namely the Stokes vector (\cite{collett2005field}):
\begin{equation}
\begin{split}
    S_0 &= \frac{1}{2} ( x_{0^\circ} + x_{45^\circ} + x_{90^\circ} + x_{135^\circ} ), \\
    S_1 &= x_{0^\circ} - x_{90^\circ}, \quad S_2 = x_{45^\circ} - x_{135^\circ},
\end{split}
\label{eq:2}
\end{equation}
where $\{ {S_0},{S_1},{S_2}\}$ denotes the Stokes vector. Accordingly, $I$, $p$, and $\phi$ in Eq. (\ref{eq:1})
can be derived from the Stokes vector:
\begin{equation}
    \begin{split}
I = \frac{{{S_0}}}{2},p = \frac{{\sqrt {S_1^2 + S_2^2} }}{{{S_0}}},\phi  = \frac{1}{2}\arctan (\frac{{S_2}}{{S_1}}).
\end{split}
    \label{eq:3}
\end{equation}
 
When polarization imaging is affected by degradations, the degraded observation ${Y} = [ y_{{0^ \circ }},y_{{{45}^ \circ }},y_{{{90}^ \circ }},y_{{{135}^ \circ }}]$ can be generally formulated as:
\begin{equation}
    \begin{split}   
Y = \mathcal{D}(X)+n,
\end{split}  
    \label{eq:4}
\end{equation}
where $n$ denotes the noise, and $\mathcal{D}$ can represent arbitrary degradation operators,
such as exposure adjustment, blur kernels (\cite{zhou2025learning}), and imaging subsampling (\cite{li2026pugdiff}).
According to Eqs. (\ref{eq:2})-(\ref{eq:4}), compared with intensity images,
$p$ and $\phi$ suffer from amplified degradation effects due to their nonlinear computation, resulting in lower image quality.
Therefore, Polarization image restoration requires not
only recovering intensity information but also
ensuring accurate estimation of polarization parameters,
which poses significant challenges for model optimization.

\begin{figure}[!t]
	\begin{center}
		\includegraphics[width=\linewidth]{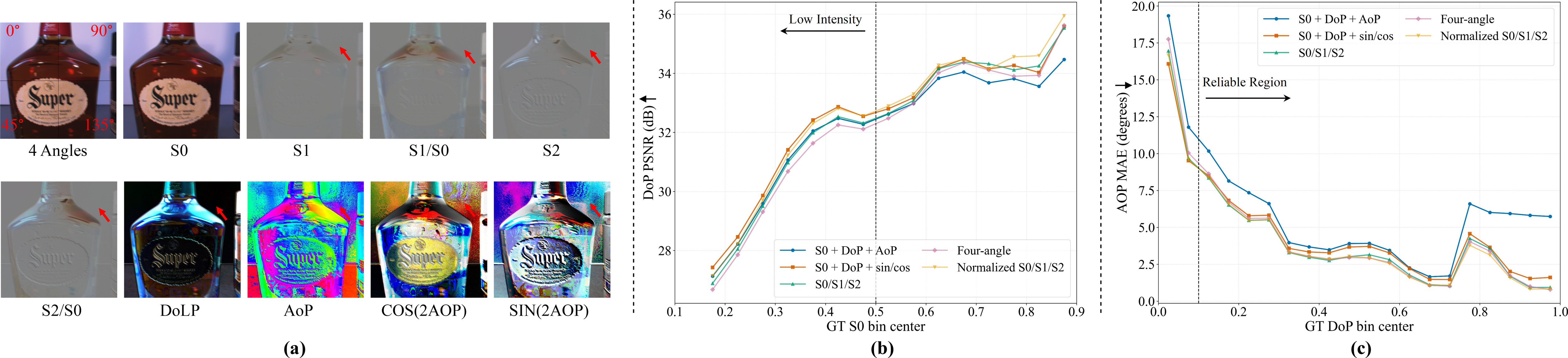}
	\end{center}
	\vspace{-1.5em}
	\caption{Representation study. (a) Representation visualization.
	(b) DoLP metric across intensity levels. Higher is better.
	(c) AoP metric across DoLP levels. Lower is better.}
	\label{fig:2}
\end{figure}

\subsection{Representation Exploration}
Before designing the network, we first address a fundamental question:
Which polarization representation (or parameterization)  is better suited
for polarization image restoration? We compare five formulations: the intensity
domain $[x_{0^\circ},x_{45^\circ},x_{90^\circ},x_{135^\circ}]$, Stokes
$[S_0,S_1,S_2]$, normalized Stokes
$[S_0,\tilde S_1,\tilde S_2]$ with
$\tilde S_i=S_i/S_0$, and the polarization-parameter domains
$[S_0,p,\phi]$ and $[S_0,p,\cos(2\phi),\sin(2\phi)]$.
Using NAFNet (\cite{chen2022simple}) as a common backbone, we evaluate all
representations on three tasks (deblurring (\cite{zhou2025learning}), LLIE (\cite{zhou2023polarization}), and demosaicking (\cite{zhou2025pidsr})) under identical settings.

As shown in Tab. \ref{tab:1}, directly restoring angular
intensities is redundant, whereas $[S_0,p,\phi]$ is difficult to optimize
because AoP is periodic. Stokes-based representations perform better, with
$[S_0,\tilde S_1,\tilde S_2]$ giving the best overall polarization accuracy
by decoupling polarization from intensity, consistent with Eq. (\ref{eq:1}). Such decoupling makes the representation
more robust to low intensity than the original Stokes
representation, as illustrated in Fig. \ref{fig:2}(b). 

Moreover, $[\tilde S_1,\tilde S_2]=[p\cos(2\phi),p\sin(2\phi)]$ provides a
continuous Cartesian encoding of AoP. Since AoP is unreliable in low-DoLP
regions (\cite{luo2026stokes}), even in the ground truth. Overfitting these
regions during optimization can degrade restoration performance. $p$ acts as a confidence weight and
reduces the influence of noise. This explains its lower reliable-region
error ($MAE_r$) than $[S_0,p,\cos(2\phi),\sin(2\phi)]$ in Tab. \ref{tab:1} and
Fig. \ref{fig:2}(c). We therefore adopt $[S_0,\tilde S_1,\tilde S_2]$ throughout.

\begin{wraptable}{r}{0.50\textwidth}
\centering
\vspace{-1.0em}
\caption{Average performance of different representations on three
restoration tasks. From (a) to (e): $[x_{0^\circ},x_{45^\circ},x_{90^\circ},x_{135^\circ}]$,
 $[S_0,S_1,S_2]$, $[S_0,\tilde S_1,\tilde S_2]$, $[S_0,p,\phi]$ and $[S_0,p,\cos(2\phi),\sin(2\phi)]$. The subscripts
 $I$ and $p$ denote intensity and DoLP, respectively. $MAE_r$ is the AOP error in reliable regions with GT DoLP $>0.1$.}
\label{tab:1}
\scriptsize
\setlength{\tabcolsep}{0.5pt}
\resizebox{\linewidth}{!}{%
\begin{tabular}{lcccccc}
\toprule
Type & \shortstack{$PSNR_I$$\uparrow$} &
\shortstack{$PSNR_p$$\uparrow$} &
\shortstack{$SSIM_I$$\uparrow$} &
\shortstack{$SSIM_p$$\uparrow$} & MAE $\downarrow$ & $MAE_r$ $\downarrow$ \\
\midrule
(a) & 30.489 & 29.390 & 0.864 & 0.713 & 16.541 & 12.973 \\
(b) & \textbf{30.603} & 29.642 & \textbf{0.864} & 0.723 & 15.740 & 12.618 \\
(c) & 30.515 & \textbf{30.110} & 0.862 & \textbf{0.727} & 15.603 & \textbf{12.101} \\
(d) & 30.118 & 29.424 & 0.857 & 0.712 & 16.982 & 13.852 \\
(e) & 30.268 & 29.790 & 0.858 & 0.722 & \textbf{14.987} & 12.503 \\
\bottomrule
\end{tabular}}
\vspace{-0.6em}
\end{wraptable}

\subsection{Composite Degradation Polarization Image Dataset}
Existing polarization datasets mainly target single degradations,
while real-world polarization imaging often suffers from coupled
degradations caused by low signal-to-noise ratio, long exposure, and DoFP sampling.
To support composite restoration, we construct PolarCDD, a large-scale polarization
dataset containing four basic degradations—low light,
motion blur, noise, and mosaic artifacts—and their combinations,
for a total of 11 degradation settings.
Blur and mosaic degradations are synthesized following (\cite{zhou2025learning}) and (\cite{luo2026polarapp}),
respectively; noise includes both synthetic and real observations,
while low-light data are captured using a DoT system.
Each degradation setting contains 50,000 training images and 100
test images\footnote{Detailed pipelines are provided in the appendix.}.

\begin{figure}[!t]
	\begin{center}
		\includegraphics[width=\linewidth]{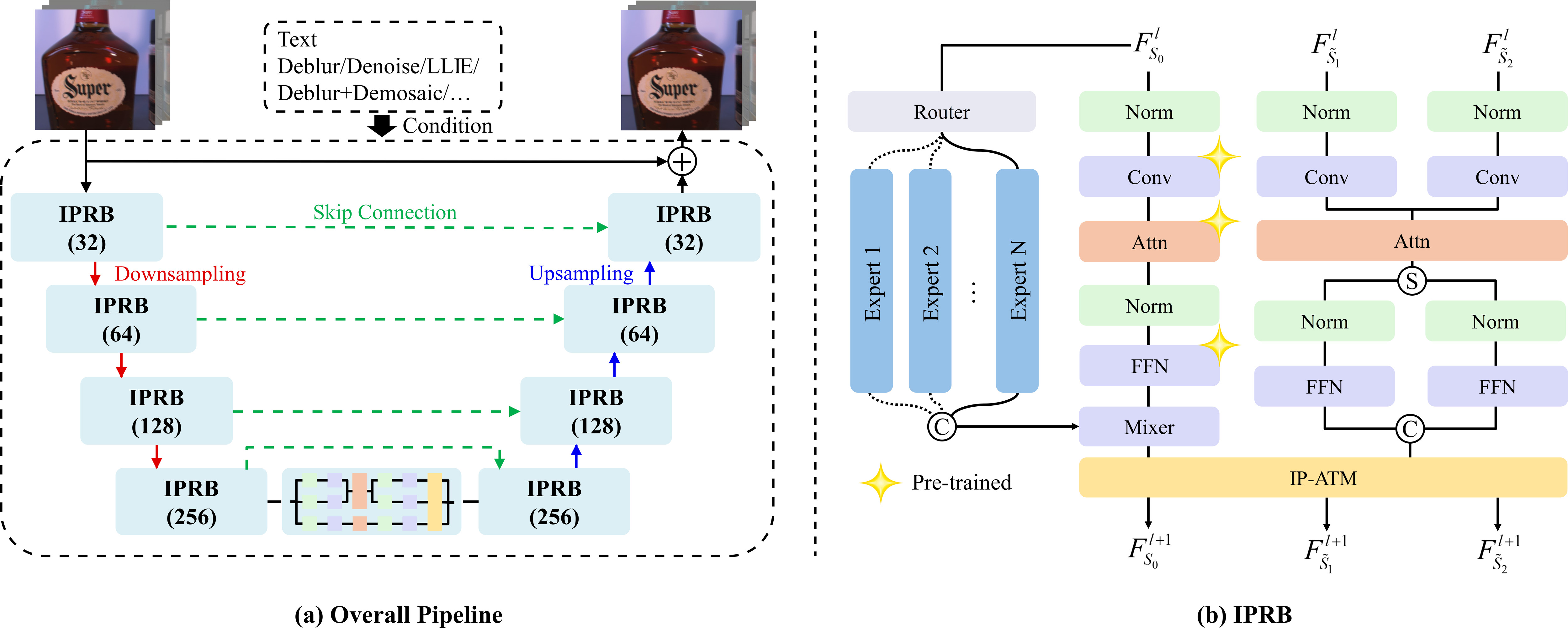}
	\end{center}
	\vspace{-1.5em}
	\caption{Overview of the proposed method.(a) The normalized
	Stokes representation is processed by
	a U-Net composed of intensity-polarization restoration blocks (IPRBs), with restoration
	behavior conditioned by textual guidance through
	instruction condition block (ICB (\cite{conde2024instructir})) . (b) Each IPRB contains intensity and
	polarization branches built upon NAFNet-style
	attention and feed-forward blocks (\cite{chen2022simple}).
	Residual connections are employed but omitted in the figure for clarity.
	The intensity branch incorporates pretrained restoration priors (\cite{conde2024instructir})
	and an MoE extension, while the polarization branch
	separately models two polarization components and
	aggregates them during attention. Finally, the intensity-polarization adaptive transfer module (IP-ATM)
	enables domain interaction and
	adapts restoration priors to the polarization domain.}
	\label{fig:3}
	\vspace{-1.5em}
\end{figure}

\subsection{Network Design}

\subsubsection{Overview}
Our network follows a U-Net architecture as shown in Fig. \ref{fig:3},
with each stage composed of several stacked Intensity-Polarization
Restoration Blocks (IPRBs). The normalized Stokes representation
$[{S_0},{\tilde S_1},{\tilde S_2}]$ are
concatenated and fed into the network.
Specifically, each IPRB adopts a parallel tri-stream design,
where three polarization representations are independently
processed by their respective backbones and interact across
domains through the Intensity-Polarization Adaptive Transfer Module (IP-ATM).
To overcome the limited distribution of existing polarization data, we introduce
general restoration priors. Since $S_0$ corresponds to a standard RGB image, pretrained
RGB restoration priors can be directly applied to its restoration.
Therefore, the IPRB backbone follows the RGB restoration model (in this work, InstructIR (\cite{conde2024instructir})
\footnote{Our design is architecture-agnostic. We
further investigate different restoration priors in the appendix.}) design, with the intensity branch
initialized from its pretrained weights. For efficiency, we reduce the
block number of the polarization branch by half, such that only half of the IPRBs
operate solely on $S_0$ in each stage. Following InstructIR, textual embeddings
are used to condition the restoration behavior of the network.

\subsubsection{IPRB}
Within each IPRB, different representations are
processed by separate branches to reduce interference
caused by distribution discrepancies across domains.
Each branch follows the standard NAFNet (\cite{chen2022simple}) design. For the intensity branch,
we further extend the backbone into an MoE architecture
to improve its capability in handling composite degradations.
Each expert is initialized with pretrained InstructIR (\cite{conde2024instructir}) weights
and mixed with the main branch through a convolutional layer.
Intensity branch employs four experts, with only the top-1 expert selected for mixture.
For the polarization branch, Stokes representations
are also processed independently, but are concatenated
before attention to enable joint aggregation within the domain.
\subsubsection{IP-ATM}

\begin{wrapfigure}{r}{0.56\textwidth}
	\centering
	\vspace{-1.2em}
	\includegraphics[width=\linewidth]{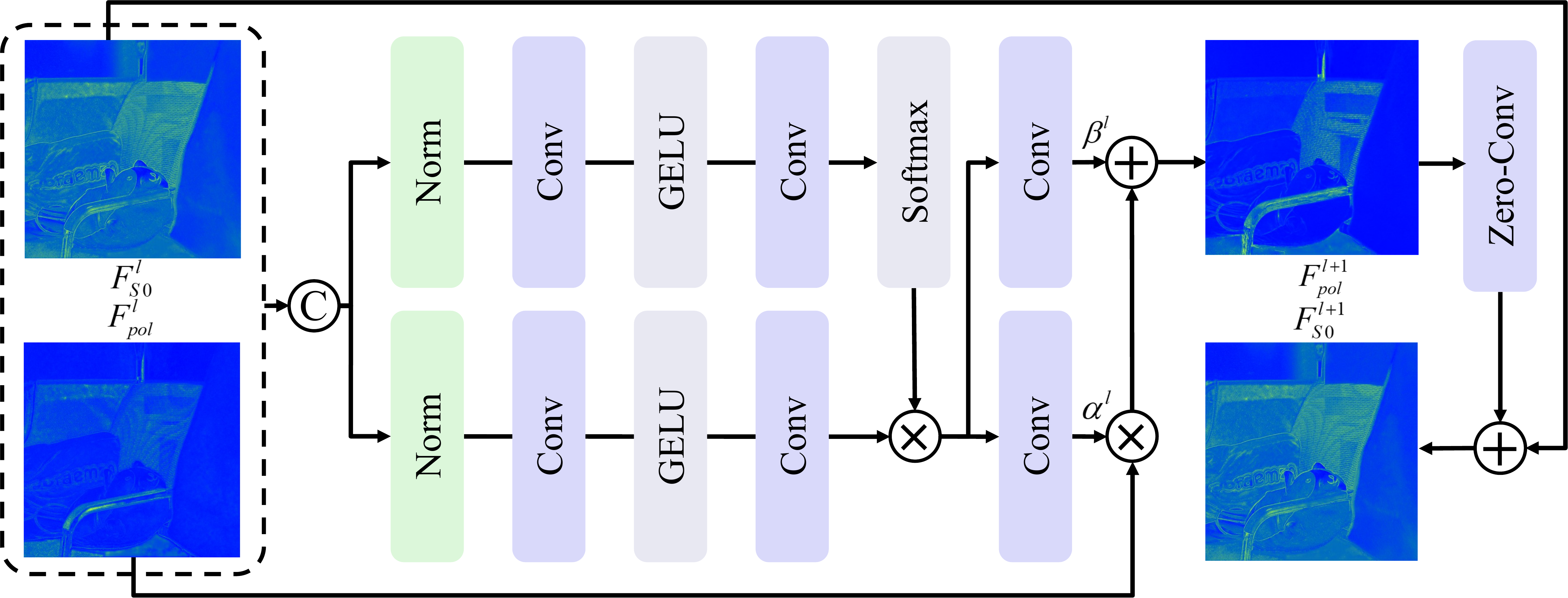}
	\caption{IP-ATM mechanism.
	Polarization representations
	selectively incorporate intensity
	restoration priors through adaptive
	scaling and shifting, followed by
	zero-convolution modulation that feeds
	polarization cues back into the intensity features.}
	\label{fig:4}
	\vspace{-0.8em}
\end{wrapfigure}

After backbone feature extraction, the
intensity and polarization features are
further complemented through a bidirectional
intensity-polarization interaction mechanism.
Our key motivation is to transfer the powerful
restoration priors learned in the intensity domain to the polarization domain.
Inspired by SFT (\cite{wang2018recovering}) and pixel AdaLN (\cite{yu2026pixeldit}), we use the
intensity features to adaptively modulate
the polarization representations through
pixel-wise scaling and shifting at the end of each layer $l$. However,
not all information encoded in the intensity
features is beneficial to polarization
restoration, and indiscriminate transfer
may introduce intensity-specific biases
into the polarization domain. We therefore
first normalize the representation $[F_{{S_0}}^l,F_{{{\tilde S}_1}}^l,F_{{{\tilde S}_2}}^l]$ to
suppress their domain-specific statistics. 
$[F_{{S_0}}^l,F_{{{\tilde S}_1}}^l,F_{{{\tilde S}_2}}^l]$ are then concatenated to predict adaptive
gating weights that selectively identify transferable features:
\begin{equation}
    \begin{split}
G^l = Softmax (Convs(F^l)), {M^l} = {G^l} \odot Convs(F^l),
\end{split}
    \label{eq:5}
\end{equation}
where $F^l$ represents the concatenated features, $Convs$ denotes convolutional
operations, and $\odot$ is the element-wise multiplication. $Softmax$ is applied along
the channel dimension. The selected features $M^l$ are further used to estimate the affine
parameters $[\alpha^l, \beta^l]$ for modulating the polarization representations:
\begin{equation}
    \begin{split}
{\alpha ^l},{\beta ^l} = Convs({M^l}),{\rm{ }}F_{pol}^{l + 1} = (1 + {\alpha ^l}) \odot F_{pol}^l + {\beta ^l},
\end{split}
    \label{eq:6}
\end{equation}
where $F_{pol}^l$ and $F_{pol}^{l + 1}$ denote the polarization features $[F_{{{\tilde S}_1}}^l,F_{{{\tilde S}_2}}^l]$ before and after modulation.

In the reverse direction, the intensity branch already has strong
restoration capability inherited from restoration priors. We therefore adopt a simple
interaction strategy,
where polarization information is injected
into the intensity features through
zero-convolution-based modulation (\cite{zhang2023adding}):
\begin{equation}
    \begin{split}
F_{{S_0}}^{l + 1} = F_{{S_0}}^l + Con{v_{zero}}(F_{Pol}^{l + 1}).
\end{split}
    \label{eq:7}
\end{equation}
The zero initialization enables
the interaction to start from an
identity-like state, preserving
the pretrained restoration behavior
at the beginning of optimization
while progressively incorporating
complementary polarization cues to
improve intensity restoration.

\begin{table}[t]
\centering
\caption{Quantitative comparison under the AIO3 setting. The superscript $*$ indicates task-specific weights. \textbf{Bold} indicates the best performance.}
\label{tab:2}
\scriptsize
\setlength{\tabcolsep}{1.5pt}

\begin{minipage}[t]{0.49\textwidth}
\centering
\begin{tabular*}{\linewidth}{@{\extracolsep{\fill}}lccccc@{}}
\toprule
Method & \shortstack{$PSNR_I$$\uparrow$} & \shortstack{$PSNR_p$$\uparrow$} &
\shortstack{$SSIM_I$$\uparrow$} & \shortstack{$SSIM_p$$\uparrow$} & MAE $\downarrow$ \\
\midrule
PolDeblur$^{\ast}$ & 29.193 & 27.232 & 0.864 & 0.694 & 15.803 \\
SSMD$^{\ast}$ & 32.601 & 29.550 & 0.909 & 0.744 & 14.416 \\
PromptIR & 25.098 & 27.554 & 0.802 & 0.690 & 15.858 \\
InstructIR & 29.128 & 28.388 & 0.873 & 0.709 & 15.596 \\
AdaIR & 25.106 & 27.516 & 0.801 & 0.690 & 15.874 \\
MoCE-IR & 31.451 & 29.012 & 0.902 & 0.729 & 15.184 \\
CDIR & 25.366 & 27.633 & 0.807 & 0.692 & 15.842 \\
SSMD & 27.880 & 27.588 & 0.842 & 0.699 & 15.495 \\
Ours & \textbf{34.029} & \textbf{29.939} & \textbf{0.933} & \textbf{0.749} & \textbf{13.991} \\
\bottomrule
\end{tabular*}
\vspace{2pt}

\textbf{(a) Deblurring dataset (\cite{zhou2025learning}). }
\end{minipage}\hfill
\begin{minipage}[t]{0.49\textwidth}
\centering
\begin{tabular*}{\linewidth}{@{\extracolsep{\fill}}lccccc@{}}
\toprule
Method & \shortstack{$PSNR_I$$\uparrow$} & \shortstack{$PSNR_p$$\uparrow$} &
\shortstack{$SSIM_I$$\uparrow$} & \shortstack{$SSIM_p$$\uparrow$} & MAE $\downarrow$ \\
\midrule
PLIE$^{\ast}$ & 38.967 & 26.876 & 0.974 & 0.689 & 16.855 \\
CPNet$^{\ast}$ & 39.868 & 28.568 & 0.980 & 0.754 & 13.954 \\
PromptIR & 38.211 & 27.856 & 0.974 & 0.727 & 14.352 \\
InstructIR & 39.171 & 28.202 & 0.976 & 0.737 & 14.405 \\
AdaIR & 37.893 & 27.541 & 0.972 & 0.720 & 14.560 \\
MoCE-IR & 39.373 & 28.758 & 0.979 & 0.745 & 14.270 \\
CDIR & 39.094 & 28.296 & 0.977 & 0.735 & 13.998 \\
SSMD & 39.549 & 27.975 & 0.977 & 0.744 & 14.059 \\
Ours & \textbf{40.512} & \textbf{29.352} & \textbf{0.982} & \textbf{0.755} & \textbf{13.340} \\
\bottomrule
\end{tabular*}
\vspace{2pt}

\textbf{(b) LLIE dataset (\cite{zhou2023polarization}). }
\end{minipage}

\vspace{8pt}

\begin{minipage}[t]{0.49\textwidth}
\centering
\begin{tabular*}{\linewidth}{@{\extracolsep{\fill}}lccccc@{}}
\toprule
Method & \shortstack{$PSNR_I$$\uparrow$} & \shortstack{$PSNR_p$$\uparrow$} &
\shortstack{$SSIM_I$$\uparrow$} & \shortstack{$SSIM_p$$\uparrow$} & MAE $\downarrow$ \\
\midrule
PIDSR$^{\ast}$ & 44.406 & 40.888 & 0.980 & 0.921 & 11.939 \\
PUGDiff$^{\ast}$ & 46.210 & 41.893 & 0.986 & 0.926 & 10.704 \\
PromptIR & 44.181 & 41.114 & 0.981 & 0.927 & 12.360 \\
InstructIR & 42.922 & 39.451 & 0.974 & 0.900 & 14.522 \\
AdaIR & 44.265 & 41.136 & 0.982 & 0.927 & 12.331 \\
MoCE-IR & 43.448 & 39.737 & 0.976 & 0.903 & 14.303 \\
CDIR & 44.188 & 40.964 & 0.981 & 0.924 & 12.027 \\
SSMD & 45.043 & 42.185 & 0.984 & 0.941 & 12.505 \\
Ours & \textbf{47.088} & \textbf{42.505} & \textbf{0.990} & \textbf{0.942} & \textbf{9.966} \\
\bottomrule
\end{tabular*}
\vspace{2pt}

\textbf{(c) Demosaicking dataset (\cite{zhou2025pidsr}). }
\end{minipage}\hfill
\begin{minipage}[t]{0.49\textwidth}
\centering
\begin{tabular*}{\linewidth}{@{\extracolsep{\fill}}lccccc@{}}
\toprule
Method & \shortstack{$PSNR_I$$\uparrow$} & \shortstack{$PSNR_p$$\uparrow$} &
\shortstack{$SSIM_I$$\uparrow$} & \shortstack{$SSIM_p$$\uparrow$} & MAE $\downarrow$ \\
\midrule
NAFNet & 30.567 & 30.697 & 0.859 & 0.747 & 15.128 \\
Restormer & 30.242 & 30.513 & 0.852 & 0.745 & 15.124 \\
PromptIR & 30.451 & 30.728 & 0.856 & 0.748 & 14.939 \\
InstructIR & 33.037 & 30.948 & 0.903 & 0.756 & 15.263 \\
AdaIR & 30.454 & 30.685 & 0.855 & 0.748 & 14.959 \\
MoCE-IR & 34.794 & 31.489 & 0.925 & 0.771 & 14.915 \\
CDIR & 30.701 & 30.780 & 0.860 & 0.749 & 14.826 \\
SSMD & 32.685 & 31.009 & 0.884 & 0.759 & 14.700 \\
Ours & \textbf{37.518} & \textbf{32.820} & \textbf{0.949} & \textbf{0.794} & \textbf{13.009} \\
\bottomrule
\end{tabular*}
\vspace{2pt}

\textbf{(d) Average performance.}
\end{minipage}
\end{table}

\begin{table}[t]
\vspace{-1 em}
\centering
\caption{Quantitative comparison on the proposed PolarCDD and
average performance. \textbf{Bold} indicates the best performance.}
\label{tab:3}
\scriptsize
\setlength{\tabcolsep}{1.5pt}

\begin{minipage}[t]{0.49\textwidth}
\centering
\begin{tabular*}{\linewidth}{@{\extracolsep{\fill}}lccccc@{}}
\toprule
Method & \shortstack{$PSNR_I$$\uparrow$} & \shortstack{$PSNR_p$$\uparrow$} &
\shortstack{$SSIM_I$$\uparrow$} & \shortstack{$SSIM_p$$\uparrow$} & MAE $\downarrow$ \\
\midrule
PromptIR & 34.248 & 30.627 & 0.919 & 0.757 & 16.914 \\
InstructIR & 33.237 & 31.309 & 0.923 & 0.779 & 17.146 \\
AdaIR & 34.729 & 31.185 & 0.925 & 0.774 & 15.687 \\
MoCE-IR & 34.542 & 32.250 & 0.938 & 0.800 & 16.615 \\
CDIR & 34.112 & 30.945 & 0.924 & 0.771 & 16.173 \\
SSMD & 34.455 & 32.031 & 0.930 & 0.804 & 15.525 \\
Ours & \textbf{37.376} & \textbf{33.191} & \textbf{0.953} & \textbf{0.826} & \textbf{13.393} \\
\bottomrule
\end{tabular*}
\vspace{2pt}

\textbf{(a) One degradation.}
\end{minipage}\hfill
\begin{minipage}[t]{0.49\textwidth}
\centering
\begin{tabular*}{\linewidth}{@{\extracolsep{\fill}}lccccc@{}}
\toprule
Method & \shortstack{$PSNR_I$$\uparrow$} & \shortstack{$PSNR_p$$\uparrow$} &
\shortstack{$SSIM_I$$\uparrow$} & \shortstack{$SSIM_p$$\uparrow$} & MAE $\downarrow$ \\
\midrule
PromptIR & 27.559 & 28.029 & 0.849 & 0.680 & 20.843 \\
InstructIR & 27.480 & 27.888 & 0.855 & 0.689 & 21.438 \\
AdaIR & 27.843 & 28.575 & 0.858 & 0.703 & 19.094 \\
MoCE-IR & 28.616 & 28.777 & 0.875 & 0.715 & 20.737 \\
CDIR & 27.486 & 28.372 & 0.854 & 0.700 & 19.957 \\
SSMD & 28.531 & 29.049 & 0.876 & 0.730 & 18.939 \\
Ours & \textbf{31.759} & \textbf{30.905} & \textbf{0.922} & \textbf{0.778} & \textbf{16.043} \\
\bottomrule
\end{tabular*}
\vspace{2pt}

\textbf{(b) Dual degradations.}
\end{minipage}

\vspace{8pt}

\begin{minipage}[t]{0.49\textwidth}
\centering
\begin{tabular*}{\linewidth}{@{\extracolsep{\fill}}lccccc@{}}
\toprule
Method & \shortstack{$PSNR_I$$\uparrow$} & \shortstack{$PSNR_p$$\uparrow$} &
\shortstack{$SSIM_I$$\uparrow$} & \shortstack{$SSIM_p$$\uparrow$} & MAE $\downarrow$ \\
\midrule
PromptIR & 24.989 & 26.688 & 0.795 & 0.647 & 20.838 \\
InstructIR & 25.279 & 26.749 & 0.807 & 0.655 & 21.247 \\
AdaIR & 24.865 & 26.940 & 0.798 & 0.659 & 19.899 \\
MoCE-IR & 26.411 & 27.260 & 0.828 & 0.673 & 20.761 \\
CDIR & 24.993 & 27.091 & 0.802 & 0.665 & 20.084 \\
SSMD & 26.053 & 27.057 & 0.831 & 0.675 & 19.656 \\
Ours & \textbf{29.904} & \textbf{29.446} & \textbf{0.901} & \textbf{0.744} & \textbf{16.397} \\
\bottomrule
\end{tabular*}
\vspace{2pt}

\textbf{(c) Triple degradations.}
\end{minipage}\hfill
\begin{minipage}[t]{0.49\textwidth}
\centering
\begin{tabular*}{\linewidth}{@{\extracolsep{\fill}}lccccc@{}}
\toprule
Method & \shortstack{$PSNR_I$$\uparrow$} & \shortstack{$PSNR_p$$\uparrow$} &
\shortstack{$SSIM_I$$\uparrow$} & \shortstack{$SSIM_p$$\uparrow$} & MAE $\downarrow$ \\
\midrule
PromptIR & 29.524 & 28.730 & 0.865 & 0.702 & 19.413 \\
InstructIR & 29.173 & 28.925 & 0.871 & 0.716 & 19.843 \\
AdaIR & 29.805 & 29.227 & 0.871 & 0.721 & 18.002 \\
MoCE-IR & 30.370 & 29.764 & 0.889 & 0.738 & 19.243 \\
CDIR & 29.442 & 29.074 & 0.870 & 0.719 & 18.604 \\
SSMD & 30.235 & 29.771 & 0.888 & 0.747 & 17.828 \\
Ours & \textbf{33.464} & \textbf{31.471} & \textbf{0.929} & \textbf{0.789} & \textbf{15.144} \\
\bottomrule
\end{tabular*}
\vspace{2pt}

\textbf{(d) Average performance.}
\end{minipage}
\vspace{-1.3 em}
\end{table}

\section{Experiment}

\subsection{Implementation Details}
We train two separate model weights.
The first is trained on a mixture of single-degradation datasets (AIO3 setting),
including PolDeblur (\cite{zhou2025learning}) for deblurring,
PLIE (\cite{zhou2023polarization}) and LLCP (\cite{xu2022colorpolarnet}) for low-light image
enhancement, and PIDSR (\cite{zhou2025pidsr}) for polarization demosaicking.
These datasets are jointly used to train a single AIO
restoration model. The second model is trained on our proposed
PolarCDD dataset. For both settings, The training and test sets are drawn from the same dataset. All models are trained for 200K iterations using the Adam (\cite{kingma2014adam}) optimizer
with an initial learning rate of $2 \times 10^{-4}$ and a cosine annealing schedule (\cite{loshchilov2016sgdr}).
The training patch size is set to $256 \times 256$ with a batch size of 8.
All experiments are conducted on NVIDIA RTX 4090 GPUs. For the loss function, we employ an $\mathcal{L}_1$
loss to constrain the intensity images, the Stokes representations
and DoP, while adopting a cosine loss for
AoP to account for its periodicity.

\begin{figure}[!t]
	\begin{center}
		\includegraphics[width=\linewidth]{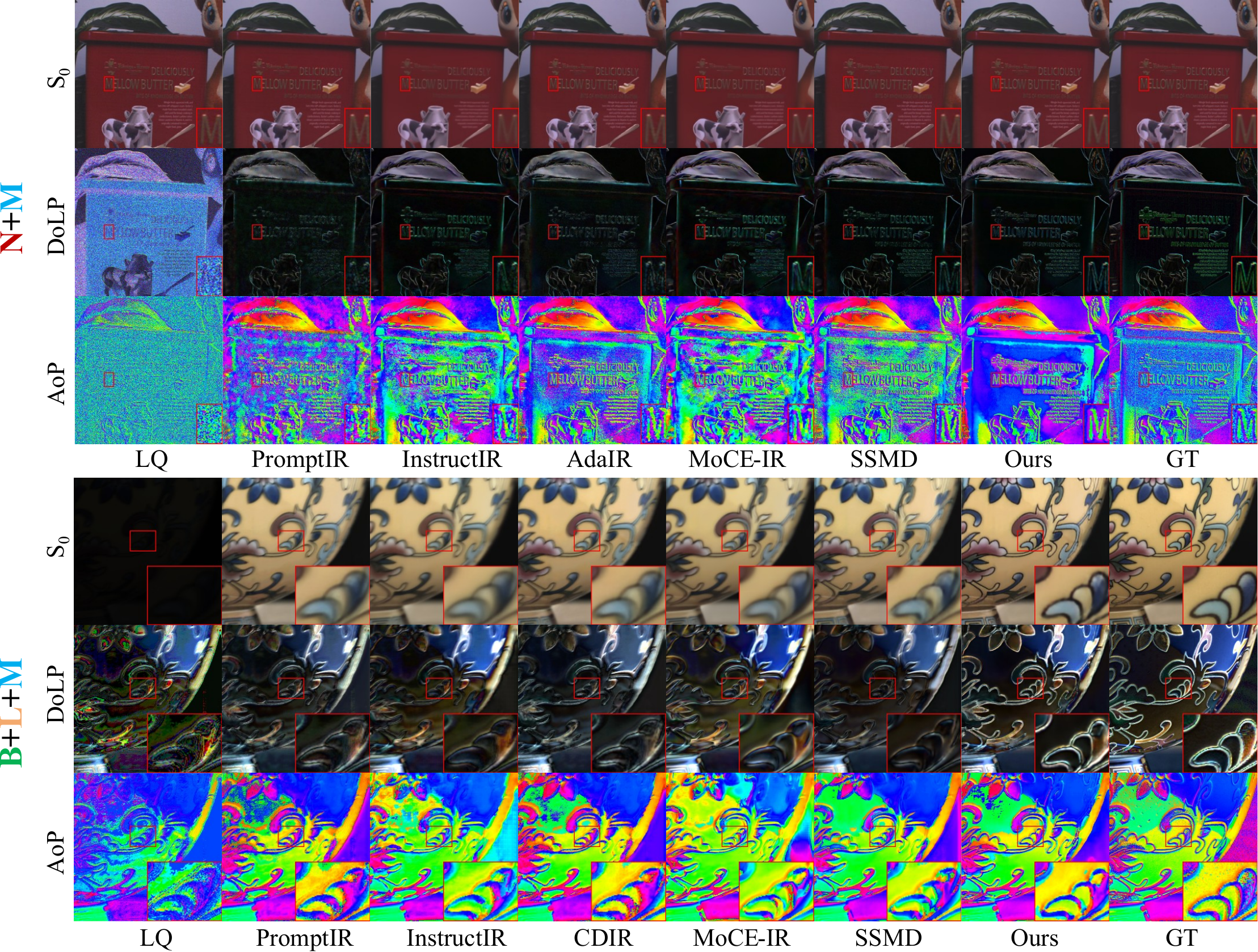}
	\end{center}
	\vspace{-0.8em}
	\caption{Visual comparisons of
	different methods on PolarCDD,
	including PromptIR (\cite{potlapalli2023promptir}), InstructIR (\cite{conde2024instructir}), AdaIR (\cite{cui2025adair}), CDIR (\cite{cui2026cdir}), MoCE-IR (\cite{zamfir2025complexity}), SSMD (\cite{zhou2026architectural}) and Ours. N is noise, B is blur,
	L is low light, and M is mosaic.}
	\label{fig:5}
	\vspace{-1 em}
\end{figure}

\begin{wraptable}{r}{0.50\textwidth}
\centering
\vspace{-2.0em}
\caption{Ablation study of the proposed framework under the AIO3 setting.}
\label{tab:4}
\scriptsize
\setlength{\tabcolsep}{1.5pt}
\resizebox{\linewidth}{!}{%
\begin{tabular}{lccccc}
\toprule
Configuration & \shortstack{$PSNR_I$$\uparrow$} &
\shortstack{$PSNR_p$$\uparrow$} &
\shortstack{$SSIM_I$$\uparrow$} &
\shortstack{$SSIM_p$$\uparrow$} & MAE $\downarrow$ \\
\midrule
No promotion                          & 36.649 & 31.125 & 0.945 & 0.757 & 14.316 \\
$+$ Polarization-to-intensity & 36.942 & 31.041 & 0.945 & 0.755 & 14.386 \\
$+$ Intensity-to-polarization (ours) & \textbf{37.518} & \textbf{32.820} & \textbf{0.949} & \textbf{0.794} & \textbf{13.009} \\
IP-ATM $\rightarrow$ Zero-Conv          & 36.365 & 31.852 & 0.940 & 0.772 & 13.620 \\
W/o gate                              & 37.449 & 32.621 & 0.947 & 0.791 & 13.100 \\
W/o MoE                               & 35.610 & 32.319 & 0.938 & 0.786 & 13.362 \\
W/o pretrained priors                  & 33.797 & 31.564 & 0.901 & 0.764 & 13.909 \\
\bottomrule
\end{tabular}}
\vspace{-0.6em}
\end{wraptable}
\vspace{-0.7em}
\subsection{Comparison with State-of-the-Art Methods}
\vspace{-0.5em}
We compare our method against two restoration paradigms.
The first category consists of RGB image restoration models,
including NAFNet (\cite{chen2022simple}), Restormer (\cite{zamir2022restormer}),
PromptIR (\cite{potlapalli2023promptir}), InstructIR (\cite{conde2024instructir}),
AdaIR (\cite{cui2025adair}), MoCE-IR (\cite{zamfir2025complexity}),
and CDIR (\cite{cui2026cdir}). To adapt these models to polarization restoration,
we consider two strategies: (1) extending the input and output channels
to match the polarization representation, i.e., the 9-channel normalized
Stokes representation used in this work, (2) or treating each
polarization angle as an individual RGB image and fine-tuning
the pretrained models in batches.
We train both variants and select the better one for each model.

The second category consists of polarization restoration methods,
including task-specific models such as PolDeblur (\cite{zhou2025learning}) for deblurring,
PLIE (\cite{zhou2023polarization}) and CPNet (\cite{xu2022colorpolarnet})
for LLIE, and PIDSR (\cite{zhou2025pidsr})
and PUGDiff (\cite{li2026pugdiff}) for polarization demosaicking.
In addition, SSMD (\cite{zhou2026architectural}) is
a general polarization restoration framework. We train SSMD
on the same mixed datasets to obtain an AIO version.

For evaluation, we use PSNR and SSIM to assess the quality of $S_0$ and DoLP,
i.e. $PSNR_{I}$, $PSNR_{p}$, $SSIM_{I}$ and $SSIM_{p}$. For AoP, we use the
mean angular error (MAE).

\textbf{Qualitative Comparisons.} Figs. \ref{fig:1} and \ref{fig:5} present
qualitative comparisons under various degradations.
Polarization properties are more vulnerable
to degradation than intensity information, while our
method produces cleaner results with sharper structural
details and fewer noise artifacts.

\textbf{Quantitative Comparisons.} Tabs. \ref{tab:2} and \ref{tab:3} report quantitative
results on public datasets and PolarCDD, respectively.
Under the AIO setting, our method substantially outperforms
general AIO models and even surpasses several task-specific methods in fidelity.
We attribute this advantage to the effective exploitation
of general restoration priors, which helps alleviate the
limitations of existing polarization training data.
In contrast, directly fine-tuning pretrained general
restoration models such as InstructIR and MoCE-IR yields
unsatisfactory performance, as these models
operate in the intensity domain and cannot adequately
capture inter-angle polarization discrepancies or
higher-level polarization representations. These results
suggest that simply transferring a pretrained RGB restoration
model is insufficient; instead, a divide-and-conquer design
that couples a general intensity restoration model with a
dedicated polarization branch is necessary to effectively
adapt general restoration priors to polarization restoration.

\begin{wrapfigure}{r}{0.56\textwidth}
	\centering
	\vspace{-2em}
	\includegraphics[width=\linewidth]{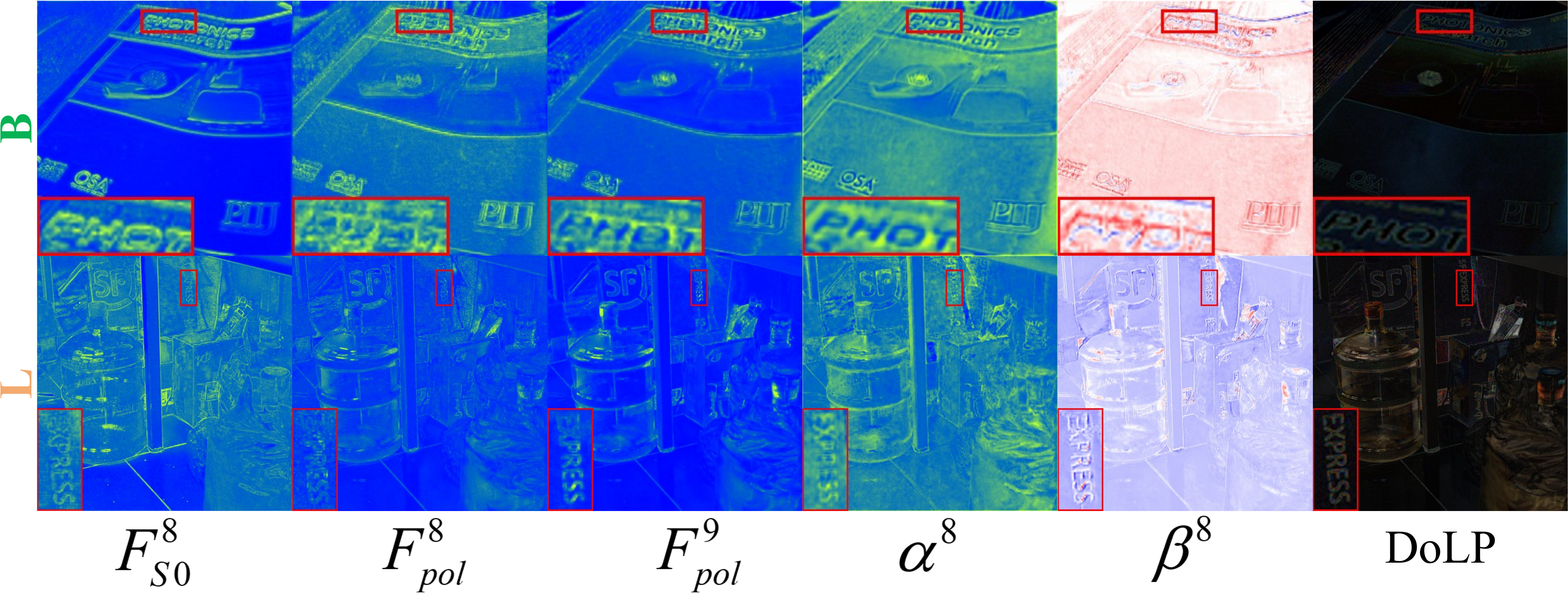}
	\vspace{-1.5em}
	\caption{Feature visualization of IP-ATM at the
	8th layer under different degradations.}
	\label{fig:6}
	\vspace{-0.8em}
\end{wrapfigure}

\subsection{Ablation Study}
\textbf{Ablation on IP-ATM.} We introduce a bidirectional
promotion mechanism to facilitate
information exchange between the intensity
and polarization domains. As shown in Tab. \ref{tab:4},
adding polarization-to-intensity modulation
improves the intensity metrics, confirming
that polarization cues provide complementary
guidance for intensity restoration. Conversely,
introducing intensity-to-polarization transfer
brings substantial gains in polarization metrics,
demonstrating the effectiveness of general restoration
priors for polarization recovery. Replacing the
intensity-to-polarization modulation with
zero-convolution leads to consistent performance
drops, suggesting that this direction requires
stronger feature modulation. Removing the gating
mechanism also degrades polarization-related metrics,
indicating that selective modulation is preferable to unrestricted feature transfer. Fig. \ref{fig:6} visualizes how intensity
features influence the polarization branch
under different degradations. After interaction,
the polarization features exhibit clearer structural
details and reduced feature noise, while
intensity-specific responses, such as the highlighted
text in the first row, are not propagated into the
polarization representation.

\textbf{Importance of General Restoration Priors.} When the general
restoration prior is removed and the intensity
branch is trained from scratch, performance drops
substantially, highlighting the importance of pretrained
restoration knowledge. Nevertheless, the polarization
metrics remain better than those obtained without
bidirectional interaction, further validating the
effectiveness of IP-ATM even in the absence of pretrained priors.
In addition, the MoE extension consistently improves
performance across all metrics and enhances robustness to
degradations outside the pretraining distribution, such as demosaicking
\footnote{Additional ablation studies are provided in the appendix.}.
\vspace{-0.7em}
\section{Conclusion}
\vspace{-0.5em}
This work presents a unified polarization
restoration network for handling diverse
degradations. We first reveal the advantages
of normalized Stokes representations for
polarization restoration, particularly under
low-intensity conditions and in reliable
polarization regions. Based on this observation,
we develop an intensity-polarization decoupled
dual-branch architecture and introduce general
restoration priors into the intensity domain to
overcome the limitations of existing polarization
data distributions. To effectively exploit these
priors for polarization restoration, we further
design an adaptive transfer mechanism that identifies
transferable intensity features and selectively
incorporates them into the polarization branch.
In addition, we establish a composite-degradation
polarization dataset to support research on
polarization restoration. Extensive
experiments on public datasets and our proposed
benchmark demonstrate the effectiveness of the proposed approach.

\FloatBarrier
\bibliography{iclr2027_conference}
\bibliographystyle{plainnat}

\appendix
\section{Appendix}
In the appendix, we first provide detailed
network architecture in \ref{A1}, followed
by the dataset construction procedure
in \ref{A2} and the implementation details
of the compared methods in \ref{A3}. \ref{A4}
analyzes model complexity, \ref{A5} investigates
the impact of different restoration priors,
and \ref{A6} examines the design of the polarization
branch. \ref{A7} introduces the loss formulation
and presents corresponding ablation studies.
\ref{A8} analyzes the mixture-of-experts mechanism, \ref{A9}
discusses the limitations of this work and potential
directions for future research, and \ref{A10} presents additional qualitative results.

\subsection{Network Architecture Details}
\label{A1}
Detailed network configurations are provided in Tab. \ref{tab:5}. The backbone of intensity-polarization restoration block (IPRB) follows the NAFNet design (\cite{chen2022simple}).
Specifically, the convolutional layers consist of $1\times1$
convolutions and $3\times3$ depth-wise convolutions, the
attention module adopts Simplified Channel Attention (SCA), and
the activation function is replaced with SimpleGate.
Residual connections are also employed but omitted in Fig. \ref{fig:3} for clarity.

\begin{table*}[t]
\centering
\caption{Configurations of the proposed network architecture. $B_0$ and
$B_{12}$ denote the numbers of NAFBlocks in the intensity ($S_0$) and
polarization ($\tilde S_1/\tilde S_2$) branches, respectively. $c_0$ and $c_{12}$ denote
their feature-channel dimensions. ICB is the instruction condition block,
IP-ATM is the intensity--polarization adaptive transfer module, and MoE uses
$E$ experts with top-$k$ routing.}
\label{tab:5}
\vspace{0.3em}
\scriptsize
\setlength{\tabcolsep}{4.0pt}
\renewcommand{\arraystretch}{1.08}
\resizebox{\textwidth}{!}{%
\begin{tabular}{llp{7.2cm}c}
\toprule
Component & Layer & Configuration & Output size \\
\midrule
Text encoder
 & Language model & BGE-Micro-v2; mean pooling; frozen & $384$ \\
 & Projection head & Linear $384\!\rightarrow\!256$; $\ell_2$ normalization & $256$ \\
\midrule
Input
 & Representation & $[S_0,\tilde S_1,\tilde S_2]$; $3+3+3$ channels & $h\times w\times9$ \\
 & Branch split & $S_0$: 3 channels; $[\tilde S_1,\tilde S_2]$: 6 channels &
 $h\times w\times(3+6)$ \\
 & Intro & $3\times3$ Conv for $S_0$; grouped $3\times3$ Conv ($g=2$) for $\tilde S_1/\tilde S_2$ &
 $h\times w\times(32+32)$ \\
\midrule
Encoder 1
 & IPRB & $(B_0,B_{12})=(2,1)$; $(c_0,c_{12})=(32,32)$; ICB; IP-ATM &
 $h\times w\times(32+32)$ \\
 & Down 1 & $2\times2$ Conv, stride 2 & $\frac h2\times\frac w2\times(64+64)$ \\
Encoder 2
 & IPRB & $(B_0,B_{12})=(2,1)$; $(c_0,c_{12})=(64,64)$; ICB; IP-ATM &
 $\frac h2\times\frac w2\times(64+64)$ \\
 & Down 2 & $2\times2$ Conv, stride 2 & $\frac h4\times\frac w4\times(128+128)$ \\
Encoder 3
 & IPRB & $(B_0,B_{12})=(4,2)$; $(c_0,c_{12})=(128,128)$; ICB; IP-ATM &
 $\frac h4\times\frac w4\times(128+128)$ \\
 & Down 3 & $2\times2$ Conv, stride 2 & $\frac h8\times\frac w8\times(256+256)$ \\
Encoder 4
 & IPRB & $(B_0,B_{12})=(8,4)$; $(c_0,c_{12})=(256,256)$; ICB; IP-ATM &
 $\frac h8\times\frac w8\times(256+256)$ \\
 & Down 4 & $2\times2$ Conv, stride 2 & $\frac h{16}\times\frac w{16}\times(512+512)$ \\
\midrule
Bottleneck
 & IPRB & $(B_0,B_{12})=(4,2)$; $(c_0,c_{12})=(512,512)$; IP-ATM &
 $\frac h{16}\times\frac w{16}\times(512+512)$ \\
\midrule
Decoder 1
 & Up 1 & $1\times1$ Conv; PixelShuffle$(2)$; skip addition &
 $\frac h8\times\frac w8\times(256+256)$ \\
 & IPRB--MoE & $(B_0,B_{12})=(2,1)$; $E=4$, top-$k=1$; ICB; IP-ATM &
 $\frac h8\times\frac w8\times(256+256)$ \\
Decoder 2
 & Up 2 & $1\times1$ Conv; PixelShuffle$(2)$; skip addition &
 $\frac h4\times\frac w4\times(128+128)$ \\
 & IPRB--MoE & $(B_0,B_{12})=(2,1)$; $E=4$, top-$k=1$; ICB; IP-ATM &
 $\frac h4\times\frac w4\times(128+128)$ \\
Decoder 3
 & Up 3 & $1\times1$ Conv; PixelShuffle$(2)$; skip addition &
 $\frac h2\times\frac w2\times(64+64)$ \\
 & IPRB--MoE & $(B_0,B_{12})=(2,1)$; $E=4$, top-$k=1$; ICB; IP-ATM &
 $\frac h2\times\frac w2\times(64+64)$ \\
Decoder 4
 & Up 4 & $1\times1$ Conv; PixelShuffle$(2)$; skip addition &
 $h\times w\times(32+32)$ \\
 & IPRB--MoE & $(B_0,B_{12})=(2,1)$; $E=4$, top-$k=1$; ICB; IP-ATM &
 $h\times w\times(32+32)$ \\
\midrule
Output
 & Ending & Separate $3\times3$ Convs; grouped Conv ($g=2$) for $\tilde S_1/\tilde S_2$; residual addition &
 $h\times w\times9$ \\
\bottomrule
\end{tabular}%
}
\end{table*}

For the mixture-of-experts (MoE) module, it is only employed in the decoder. Each expert shares the same architecture as the intensity branch. Following MoCE-IR (\cite{zamfir2025complexity}), the router uses a linear layer
to map the input intensity features and text embeddings into expert
selection probabilities. The Mixer is implemented as a
convolutional layer to fuse the outputs of the selected
experts with the intensity-branch features.

For the polarization branch, the two polarization
representations are temporarily aggregated during
the attention operation and then split back into
separate features for independent processing.
They are finally merged again to facilitate information transfer.

Following InstructIR (\cite{conde2024instructir}),
we use textual guidance to control the restoration behavior.
The input text is encoded into an embedding
by a text encoder (\cite{reimers2019sentence}) and injected into each IPRB through instruction condition block (ICB).
Notably, ICB modulates only the intensity features, while
the corresponding restoration mode is propagated to the
polarization domain through intensity-polarization adaptive transfer module (IP-ATM). For each degradation setting,
we use a fixed text prompt.
For example, the prompt is “deblur” for motion deblurring
and “denoise+demosaic” for the joint noise-and-mosaic degradation.
This fixed-prompt design allows the text embeddings to be
precomputed offline, enabling the text encoder to be removed
during inference for improved efficiency.

\subsection{PolarCDD Dataset Construction}
\label{A2}
\begin{figure}[!t]
	\begin{center}
		\includegraphics[width=\linewidth]{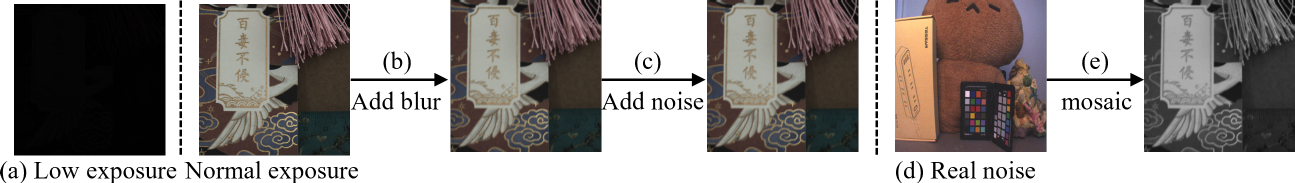}
	\end{center}
	\vspace{-1em}
	\caption{Degradation synthesis pipeline. low light: (a), blur: (b),
	noise: (c) or (d), mosaic: (e), low light+blur: (a)+(b), low light+mosaic: (a)+(e),
	blur+noise: (b)+(c), blur+mosaic: (b)+(e), noise+mosaic:
	(c)+(e) or (d)+(e), low light+blur+mosaic: (a)+(b)+(e),
	blur+noise+mosaic: (b)+(c)+(e).}
	\label{fig:7}
\end{figure}

The synthesis pipeline of PolarCDD is illustrated in Fig. \ref{fig:7}.
To isolate the effect of each degradation,
particularly mosaic artifacts, our data
are primarily collected using DoT imaging systems
(\cite{morimatsu2020monochrome}, \cite{qiu2021linear},
\cite{li2025demosaicking}, \cite{zhou2025pidsr}, and
\cite{rahman2025polarization}).
Most images in these datasets have resolutions above
$1024\times1024$, allowing us to augment them through
cropping. We use $256\times256$ crops for training
and $512\times512$ crops for testing,
while ensuring that the training and test sets
contain no images from the same scene.
Below, we detail the construction of each degradation type.

\textbf{Motion blur.} 
Due to the limited frame rate of polarization cameras,
collecting paired sharp and blurred polarization images
is challenging. We therefore follow the data synthesis
strategy of PolDeblur \cite{zhou2025learning} to generate training pairs.
The procedure is summarized as follows: (1) We first generate random camera
motion trajectories. (2) The spatially carying blur model is applied to convert
the motion trajectories into pixel-wise motion paths. (3) Each sharp polarization angles
is warped along the corresponding pixel-wise trajectories to generate a sequence of latent
sharp frames. (4) These latent frames are averaged to synthesized blurry images.

\textbf{Low light.} We capture multiple
scenes at different exposure levels using a
DoT system. Based on the exposure time, the
images are categorized into four levels: extremely
low light, low light, normal exposure, and overexposure.

\textbf{Noise.} Noise data are obtained
in two ways. One is synthetically generated
using Gaussian noise (\cite{zhou2025learning}), while the other is
collected from a real-noise dataset (\cite{rahman2025polarization}).
The real noise is further divided into three levels. 

\textbf{Mosaic artifacts.} We synthesize mosaic
observations from DoT images using the forward
imaging operator (\cite{luo2026polarapp}) to construct training pairs.

\textbf{Composite Degradations.} We first select
normally exposed or low-light scenes according
to the exposure level, followed by blur synthesis,
noise addition, and finally mosaic generation.
When real-noise images are used, we do not apply
additional blur or noise, as these operations
may alter their inherent noise statistics.
Similarly, no extra noise is added to low-light scenes,
since low-light acquisition already introduces substantial noise.
Finally, PolarCDD contains 11 degradation settings: low light, blur,
noise, mosaic, low light+blur, low light+mosaic,
blur+noise, blur+mosaic, noise+mosaic, low light+blur+mosaic,
and blur+noise+mosaic.

\subsection{Compared Methods Implementation Details}
\label{A3}
In this section, we detail
the specific configurations of
each comparison method. We reproduce each
method following its official implementation or the
configuration described in the corresponding paper. Given their
different computational complexities,
we aim to keep the training time and
GPU memory consumption comparable across methods.

\textbf{NAFNet.} An activation-free NAFNet
using 9-channel normalized Stokes
input. It has width=32, four encoder levels with [1,1,1,28] blocks,
one middle block, and [1,1,1,1] decoder blocks (\cite{chen2022simple}).

\textbf{Restormer.} A four-level
Restormer using MDTA and GDFN
blocks with dim=32, block numbers
[4,6,6,8], heads [1,2,4,8], and four
refinement blocks. It takes 9-channel
normalized Stokes input. (\cite{zamir2022restormer})

\textbf{PromptIR.} A Restormer-based model
with learnable degradation prompts
inserted at three decoder scales.
It uses dim=48, blocks [4,6,6,8], heads [1,2,4,8],
four refinement blocks, and 9-channel input/output (\cite{potlapalli2023promptir}). 

\textbf{InstructIR.} A NAFNet-based
instruction-conditioned model with
ICB modules after each encoder and decoder level.
It uses width=32, encoder blocks [2,2,4,8],
four middle blocks, decoder blocks [2,2,2,2],
and a 256-dimensional text embedding.
Four RGB polarization angles are processed independently (\cite{conde2024instructir}).

\textbf{AdaIR.} A Restormer-based
architecture with adaptive frequency
modules in the bottleneck and decoder.
It uses dim=48, blocks [4,6,6,8], heads
[1,2,4,8], four refinement blocks, and
9-channel input/output (\cite{cui2025adair}). 

\textbf{MoCE-IR.} A Restormer-style
encoder with shared decoder branches
and frequency-domain mixture-of-experts
adapters. It uses dim=48, encoder blocks
[4,6,6,8], decoder blocks [2,4,4], four
experts, and top-k=1 routing. Four RGB
angles are processed independently (\cite{zamfir2025complexity}).

\textbf{CDIR.} A dynamic low-rank
attention network using rank-8 DLA
and multi-scale feed-forward blocks.
It has width=32, encoder depths [2,3,6],
six latent blocks, and decoder depths [3,6].
The input and output contain nine normalized
Stokes channels (\cite{cui2026cdir}). 

\textbf{SSMD.} A Transformer-based model that jointly processes
four-angle images and Stokes features.
It uses dim=32,
six CDCI stages with [4,6,6,6,4,4]
units and heads [1,2,4,2,1,1].
The input and output contain 12 four-angle channels following
their original paper (\cite{zhou2026architectural}).

\subsection{Complexity Analysis}
\label{A4}

Tab. \ref{tab:6} compares the computational
complexity of different methods.
Owing to its dual-branch architecture
and MoE design, our model has the largest
parameter count. Nevertheless, its
convolution-dominated backbone maintains
competitive memory consumption, computational cost,
and inference speed. In contrast,
many restoration models are Transformer-based.
Despite having relatively few parameters, such as
Restormer (\cite{zamir2022restormer}) and SSMD (\cite{zhou2026architectural}), their attention operations
introduce substantial computational and memory overhead,
making them less amenable to efficient scaling.

\begin{table}[t]
\centering
\caption{Complexity comparison. ``9ch'' denotes evaluation with a
nine-channel normalized Stokes input adaptation. GPU memory, FLOPs, and latency are measured
on $512 \times 512$ images using an NVIDIA RTX 4090 GPU.}
\label{tab:6}
\vspace{0.3em}
\small
\setlength{\tabcolsep}{10pt}
\begin{tabular}{lcccc}
\toprule
Method & Params (M) & GPU memory (GB) & FLOPs (T) & Latency (ms) \\
\midrule
NAFNet          & 17.112 & 1.068 & 0.294 &  82.91 \\
NAFNet-9ch     & 17.115 & 0.321 & 0.074 &  29.14 \\
Restormer       & 11.740 & 5.998 & 1.029 & 684.08 \\
Restormer-9ch  & 11.745 & 1.533 & 0.259 & 175.47 \\
PromptIR        & 35.377 & 9.179 & 2.530 & 915.41 \\
PromptIR-9ch   & 35.385 & 2.474 & 0.635 & 228.93 \\
InstructIR      & 33.334 & 1.353 & 0.259 & 129.70 \\
InstructIR-9ch & 33.337 & 0.430 & 0.066 &  29.59 \\
AdaIR           & 28.785 & 9.232 & 2.355 & 977.52 \\
AdaIR-9ch      & 28.865 & 2.495 & 0.591 & 255.68 \\
MoCE-IR         & 25.354 & 4.101 & 1.591 & 917.63 \\
MoCE-IR-9ch    & 25.359 & 1.226 & 0.355 & 197.97 \\
CDIR            &  3.401 & 1.859 & 0.280 & 199.87 \\
CDIR-9ch       &  3.409 & 0.484 & 0.072 &  49.02 \\
SSMD            &  4.794 & 0.916 & 0.240 & 156.47 \\
Ours            & 43.427 & 0.903 & 0.185 &  86.05 \\
\bottomrule
\end{tabular}
\end{table}

\subsection{Analysis of Restoration Priors}
\label{A5}

\begin{wrapfigure}{r}{0.56\textwidth}
	\centering
	\vspace{-1em}
	\includegraphics[width=\linewidth]{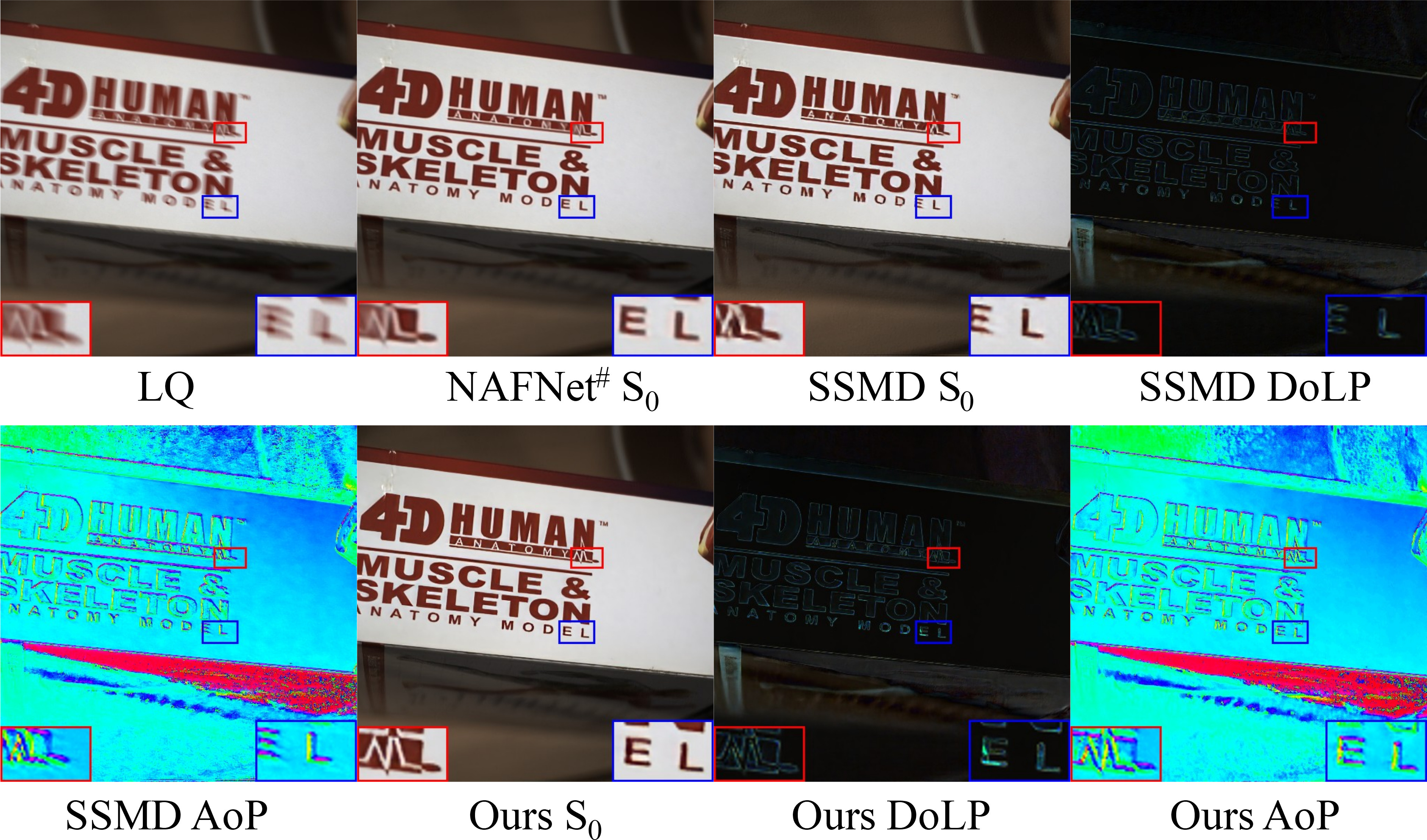}
	\vspace{-2em}
	\caption{Visual comparison on real-world motion blur.
	The superscript $\#$ denotes the pretrained model.}
	\label{fig:14}
	\vspace{-0.8em}
\end{wrapfigure}

Fig. \ref{fig:14} demonstrates the advantage of
pretrained restoration priors over
polarization-specific models. For
motion deblurring, realistic blur is
difficult to synthesize from polarization
data to the same extent as in general image
restoration. As a result, the pretrained restoration
model (NAFNet (\cite{chen2022simple})) produce better visual results on real-world
polarization data than the expert model trained solely
on polarization datasets. The polarization-specific expert
fail to produce visually satisfactory results or
restore blurred pixels to their correct spatial locations.
Our method further transfers
these strong restoration priors into the polarization
domain, yielding substantial improvements under real-world
degradations.

Our method combines a general restoration model
with a dedicated polarization branch to transfer
restoration priors into the polarization domain.
This design is architecture-agnostic,
allowing us to further investigate the effect
of different pretrained priors. As shown in Tab. \ref{tab:7},
the 3-channel fine-tuned models generally outperform
their 9-channel polarization-aware counterparts,
highlighting the importance of pretrained restoration
priors in overcoming the limitations of polarization-specific
training. Moreover, our dual-branch design consistently
achieves better performance when built upon NAFNet (\cite{chen2022simple}),
Restormer (\cite{zamir2022restormer}), MoCE-IR (\cite{zamfir2025complexity}), and InstructIR (\cite{conde2024instructir}). This not only
validates the necessity of separately modeling
intensity and polarization features, but also
demonstrates that our framework can effectively
adapt intensity-domain restoration priors to
polarization restoration.

We ultimately select InstructIR as the
restoration prior for two reasons. First,
it provides strong restoration performance
and generalization, outperforming MoCE-IR on
polarization data. Second, its lightweight
backbone is easier to extend than Transformer-based
architectures, allowing us to incorporate the
polarization branch with relatively limited
computational overhead.

\begin{table}[h!]
\centering
\caption{Comparison of different input and branch designs with general
restoration priors. ``9ch'' directly processes the nine-channel input,
``3ch'' uses the three-channel intensity representation, and ``Ours'' uses
our intensity--polarization dual-branch design. We report the
performance of all experiments at 100k training iterations. \textbf{Bold} indicates the
best performance in each subtable.}
\label{tab:7}
\fontsize{6.5pt}{7.5pt}\selectfont
\setlength{\tabcolsep}{1.5pt}

\begin{minipage}[t]{0.49\textwidth}
\centering
\begin{tabular*}{\linewidth}{@{\extracolsep{\fill}}lccccc@{}}
\toprule
Configuration & \shortstack{$PSNR_I$$\uparrow$} & \shortstack{$PSNR_p$$\uparrow$} &
\shortstack{$SSIM_I$$\uparrow$} & \shortstack{$SSIM_p$$\uparrow$} & MAE $\downarrow$ \\
\midrule
NAFNet-9ch     & 25.788 & 27.869 & 0.818 & 0.696 & 15.948 \\
NAFNet     & 33.180 & 28.554 & 0.914 & 0.728 & 15.566 \\
NAFNet-Ours    & 35.468 & 30.626 & 0.944 & 0.771 & 13.577 \\
Restormer-9ch  & 33.307 & 29.536 & 0.923 & 0.743 & 14.281 \\
Restormer  & 32.299 & 29.190 & 0.903 & 0.736 & 15.069 \\
Restormer-Ours & \textbf{36.574} & \textbf{31.014} & \textbf{0.951} &
\textbf{0.780} & \textbf{13.246} \\
\bottomrule
\end{tabular*}
\vspace{2pt}

\textbf{(a) Deblurring performance on PolDeblur dataset.}
\end{minipage}\hfill
\begin{minipage}[t]{0.49\textwidth}
\centering
\begin{tabular*}{\linewidth}{@{\extracolsep{\fill}}lccccc@{}}
\toprule
Configuration & \shortstack{$PSNR_I$$\uparrow$} & \shortstack{$PSNR_p$$\uparrow$} &
\shortstack{$SSIM_I$$\uparrow$} & \shortstack{$SSIM_p$$\uparrow$} & MAE $\downarrow$ \\
\midrule
MoCE-IR-9ch      & 28.608 & 26.396 & 0.812 & 0.605 & 22.433 \\
MoCE-IR      & 29.483 & 28.627 & 0.862 & 0.708 & 20.645 \\
MoCE-IR-Ours     & 31.306 & 29.332 & 0.890 & 0.732 & 18.160 \\
InstructIR-9ch   & 28.799 & 26.483 & 0.816 & 0.594 & 22.624 \\
InstructIR  & 29.058 & 28.315 & 0.865 & 0.703 & 20.536 \\
InstructIR-Ours  & \textbf{31.832} & \textbf{29.932} & \textbf{0.899} &
\textbf{0.752} & \textbf{16.791} \\
\bottomrule
\end{tabular*}
\vspace{2pt}

\textbf{(b) MoCE-IR and InstructIR restoration priors at PolarCDD settings.}
\end{minipage}
\end{table}

\subsection{Polarization Branch Design}
\label{A6}
As shown in Tab. \ref{tab:5}, the polarization branch uses
only half as many blocks as the intensity branch,
as we find that this lightweight design already
achieves strong performance. According to Tab. \ref{tab:8},
halving the polarization branch leads to only a 0.1 dB
drop in $PSNR_p$, while the MAE remains nearly
unchanged. We further evaluate a tiny branch,
where the polarization branch is deployed only
at Encoder 1/2, Decoder 1/2, and the bottleneck.
Its performance is consistently lower than the
half configuration, indicating the importance
of polarization processing across all feature levels.

To further examine the role of restoration priors,
we remove IP-ATM from the full polarization branch.
This results in a substantial performance drop,
falling well below both the half and tiny variants.
These results suggest that simply increasing the depth
of the polarization branch is less effective than
introducing restoration priors for
overcoming the performance bottleneck.

\begin{table}[h!]
\centering
\caption{Ablation study on the polarization branch and feature promotion under the AIO3 setting.
We report the performance of all experiments at 100k training iterations. \textbf{Bold} indicates the best performance.}
\label{tab:8}
\small
\setlength{\tabcolsep}{8pt}
\begin{tabular}{lccccc}
\toprule
Configuration & \shortstack{$PSNR_I$$\uparrow$} & \shortstack{$PSNR_p$$\uparrow$} &
\shortstack{$SSIM_I$$\uparrow$} & \shortstack{$SSIM_p$$\uparrow$} & MAE $\downarrow$ \\
\midrule
Full polarization branch
 & \textbf{36.034} & \textbf{32.342} & 0.934 & \textbf{0.784} & \textbf{13.325} \\
Full branch w/o IP-ATM
 & 35.141 & 31.119 & 0.929 & 0.756 & 14.323 \\
Tiny polarization branch
 & 35.744 & 32.017 & 0.934 & 0.777 & 13.573 \\
Half polarization branch (Ours)
 & 35.893 & 32.261 & \textbf{0.934} & 0.782 & 13.358 \\
\bottomrule
\end{tabular}
\end{table}

\subsection{Loss Function Ablation}
\label{A7}

Our loss function is formulated as follows:

\begin{equation}
\begin{array}{l}
{\mathcal L_{ang}} = {\left\| {\hat X - X} \right\|_1},{\rm{ }}{\mathcal L_{stokes}} = {\left\| {\hat S - S} \right\|_1},{\rm{ }}{\mathcal L_p} = {\left\| {\hat p - p} \right\|_1},\\
{\rm{ }}{\mathcal L_\phi } = {\left\| {\cos (2\hat \phi ) - \cos (2\phi )} \right\|_1} + {\left\| {\sin (2\hat \phi ) - \sin (2\phi )} \right\|_1},\\
\mathcal L = {\mathcal L_{ang}} + {\lambda _1}{\mathcal L_{stokes}} + {\lambda _2}{\mathcal L_p} + {\lambda _3}{\mathcal L_\phi },
\end{array}
    \label{eq:8}
\end{equation}
where ${\left\|  \cdot  \right\|_1}$ denotes the L1 norm. The GT ${X} = [ x_{{0^ \circ }},x_{{{45}^ \circ }},x_{{{90}^ \circ }},x_{{{135}^ \circ }}]$,
$S=[S_0,S_1,S_2]$. $\lambda _1=0.5$, $\lambda _2=0.08$,
and $\lambda _3=0.04$ are set to keep the different loss
terms on comparable scales. The hat notation denotes network
predictions. Although the network directly predicts
the normalized Stokes representation,
the corresponding intensity and polarization parameters
can be analytically derived from it,
allowing us to impose supervision in both
the intensity and polarization parameter domains.

Tab. \ref{tab:9} reports the performance of
different loss combinations.
Progressively introducing
$\mathcal L_{stokes}$ and $\mathcal L_{p,\phi}$ consistently
improves the polarization-related
metrics while maintaining strong
intensity performance. In contrast,
$\mathcal L_{ang}$ is critical for intensity
restoration, and removing it leads
to a substantial degradation in the
intensity metrics. Based on these
observations, we adopt the current loss configuration.

\begin{table}[h!]
\centering
\caption{Ablation study on different loss-function configurations under the AIO3 settings.
We report the performance of all experiments at 100k training iterations. \textbf{Bold} indicates the best performance.}
\label{tab:9}
\small
\setlength{\tabcolsep}{8pt}
\begin{tabular}{lccccc}
\toprule
Configuration & \shortstack{$PSNR_I$$\uparrow$} & \shortstack{$PSNR_p$$\uparrow$} &
\shortstack{$SSIM_I$$\uparrow$} & \shortstack{$SSIM_p$$\uparrow$} & MAE $\downarrow$ \\
\midrule
$\mathcal L_{ang}$
 & \textbf{36.259} & 31.224 & \textbf{0.937} & 0.757 & 14.986 \\
$+$ $\mathcal L_{stokes}$
 & 36.129 & 31.543 & 0.936 & 0.766 & 14.518 \\
$+$ $\mathcal L_{p,\phi}$ (Ours)
 & 35.893 & \textbf{32.261} & 0.934 & \textbf{0.782} & 13.358 \\
w/o $\mathcal L_{ang}$
 & 35.120 & 32.211 & 0.929 & 0.782 & \textbf{13.357} \\
\bottomrule
\end{tabular}
\end{table}

\begin{table}[t]
\centering
\caption{Comparison of Ours and Ours w/o MoE under the AIO3 setting. \textbf{Bold} indicates the best performance.}
\label{tab:10}
\scriptsize
\setlength{\tabcolsep}{0.5pt}

\begin{minipage}[t]{0.49\textwidth}
\centering
\begin{tabular*}{\linewidth}{@{\extracolsep{\fill}}lccccc@{}}
\toprule
Method & \shortstack{$PSNR_I$$\uparrow$} & \shortstack{$PSNR_p$$\uparrow$} &
\shortstack{$SSIM_I$$\uparrow$} & \shortstack{$SSIM_p$$\uparrow$} & MAE $\downarrow$ \\
\midrule
Ours-noMoE & 32.032 & 29.470 & 0.919 & 0.739 & 14.241 \\
Ours & \textbf{34.029} & \textbf{29.939} & \textbf{0.933} & \textbf{0.749} & \textbf{13.991} \\
\bottomrule
\end{tabular*}
\vspace{2pt}

\textbf{(a) Deblurring dataset (\cite{zhou2025learning}). }
\end{minipage}\hfill
\begin{minipage}[t]{0.49\textwidth}
\centering
\begin{tabular*}{\linewidth}{@{\extracolsep{\fill}}lccccc@{}}
\toprule
Method & \shortstack{$PSNR_I$$\uparrow$} & \shortstack{$PSNR_p$$\uparrow$} &
\shortstack{$SSIM_I$$\uparrow$} & \shortstack{$SSIM_p$$\uparrow$} & MAE $\downarrow$ \\
\midrule
Ours-noMoE & \textbf{40.582} & 29.044 & \textbf{0.982} & 0.753 & 13.380 \\
Ours & 40.512 & \textbf{29.352} & 0.982 & \textbf{0.755} & \textbf{13.340} \\
\bottomrule
\end{tabular*}
\vspace{2pt}

\textbf{(b) LLIE dataset (\cite{zhou2023polarization}). }
\end{minipage}

\vspace{8pt}

\begin{minipage}[t]{0.49\textwidth}
\centering
\begin{tabular*}{\linewidth}{@{\extracolsep{\fill}}lccccc@{}}
\toprule
Method & \shortstack{$PSNR_I$$\uparrow$} & \shortstack{$PSNR_p$$\uparrow$} &
\shortstack{$SSIM_I$$\uparrow$} & \shortstack{$SSIM_p$$\uparrow$} & MAE $\downarrow$ \\
\midrule
Ours-noMoE & 44.852 & 41.846 & 0.983 & 0.936 & 10.717 \\
Ours & \textbf{47.088} & \textbf{42.505} & \textbf{0.990} & \textbf{0.942} & \textbf{9.966} \\
\bottomrule
\end{tabular*}
\vspace{2pt}

\textbf{(c) Demosaicking dataset (\cite{zhou2025pidsr}). }
\end{minipage}\hfill
\begin{minipage}[t]{0.49\textwidth}
\centering
\begin{tabular*}{\linewidth}{@{\extracolsep{\fill}}lccccc@{}}
\toprule
Method & \shortstack{$PSNR_I$$\uparrow$} & \shortstack{$PSNR_p$$\uparrow$} &
\shortstack{$SSIM_I$$\uparrow$} & \shortstack{$SSIM_p$$\uparrow$} & MAE $\downarrow$ \\
\midrule
Ours-noMoE & 35.610 & 32.319 & 0.938 & 0.786 & 13.362 \\
Ours & \textbf{37.518} & \textbf{32.820} & \textbf{0.949} & \textbf{0.794} & \textbf{13.009} \\
\bottomrule
\end{tabular*}
\vspace{2pt}

\textbf{(d) Average performance.}
\end{minipage}
\vspace{10pt}

\centering
\caption{Comparison of Ours and Ours w/o MoE on the PolarCDD dataset. \textbf{Bold} indicates the best performance.}
\label{tab:11}
\scriptsize
\setlength{\tabcolsep}{0.5pt}

\begin{minipage}[t]{0.49\textwidth}
\centering
\begin{tabular*}{\linewidth}{@{\extracolsep{\fill}}lccccc@{}}
\toprule
Method & \shortstack{$PSNR_I$$\uparrow$} & \shortstack{$PSNR_p$$\uparrow$} &
\shortstack{$SSIM_I$$\uparrow$} & \shortstack{$SSIM_p$$\uparrow$} & MAE $\downarrow$ \\
\midrule
Ours-noMoE & 37.143 & 33.075 & 0.951 & \textbf{0.839} & 14.024 \\
Ours & \textbf{37.376} & \textbf{33.191} & \textbf{0.953} & 0.826 & \textbf{13.393} \\
\bottomrule
\end{tabular*}
\vspace{2pt}

\textbf{(a) One degradation.}
\end{minipage}\hfill
\begin{minipage}[t]{0.49\textwidth}
\centering
\begin{tabular*}{\linewidth}{@{\extracolsep{\fill}}lccccc@{}}
\toprule
Method & \shortstack{$PSNR_I$$\uparrow$} & \shortstack{$PSNR_p$$\uparrow$} &
\shortstack{$SSIM_I$$\uparrow$} & \shortstack{$SSIM_p$$\uparrow$} & MAE $\downarrow$ \\
\midrule
Ours-noMoE & \textbf{31.795} & 30.411 & 0.920 & 0.772 & 16.915 \\
Ours & 31.759 & \textbf{30.905} & \textbf{0.922} & \textbf{0.778} & \textbf{16.043} \\
\bottomrule
\end{tabular*}
\vspace{2pt}

\textbf{(b) Dual degradations.}
\end{minipage}

\vspace{8pt}

\begin{minipage}[t]{0.49\textwidth}
\centering
\begin{tabular*}{\linewidth}{@{\extracolsep{\fill}}lccccc@{}}
\toprule
Method & \shortstack{$PSNR_I$$\uparrow$} & \shortstack{$PSNR_p$$\uparrow$} &
\shortstack{$SSIM_I$$\uparrow$} & \shortstack{$SSIM_p$$\uparrow$} & MAE $\downarrow$ \\
\midrule
Ours-noMoE & 29.655 & 28.871 & 0.898 & 0.728 & 17.178 \\
Ours & \textbf{29.904} & \textbf{29.446} & \textbf{0.901} & \textbf{0.744} & \textbf{16.397} \\
\bottomrule
\end{tabular*}
\vspace{2pt}

\textbf{(c) Triple degradations.}
\end{minipage}\hfill
\begin{minipage}[t]{0.49\textwidth}
\centering
\begin{tabular*}{\linewidth}{@{\extracolsep{\fill}}lccccc@{}}
\toprule
Method & \shortstack{$PSNR_I$$\uparrow$} & \shortstack{$PSNR_p$$\uparrow$} &
\shortstack{$SSIM_I$$\uparrow$} & \shortstack{$SSIM_p$$\uparrow$} & MAE $\downarrow$ \\
\midrule
Ours-noMoE & 33.351 & 31.100 & 0.927 & 0.788 & 15.912 \\
Ours & \textbf{33.464} & \textbf{31.471} & \textbf{0.929} & \textbf{0.789} & \textbf{15.144} \\
\bottomrule
\end{tabular*}
\vspace{2pt}

\textbf{(d) Average performance.}
\end{minipage}
\end{table}

\begin{figure}[!t]
	\begin{center}
		\includegraphics[width=\linewidth]{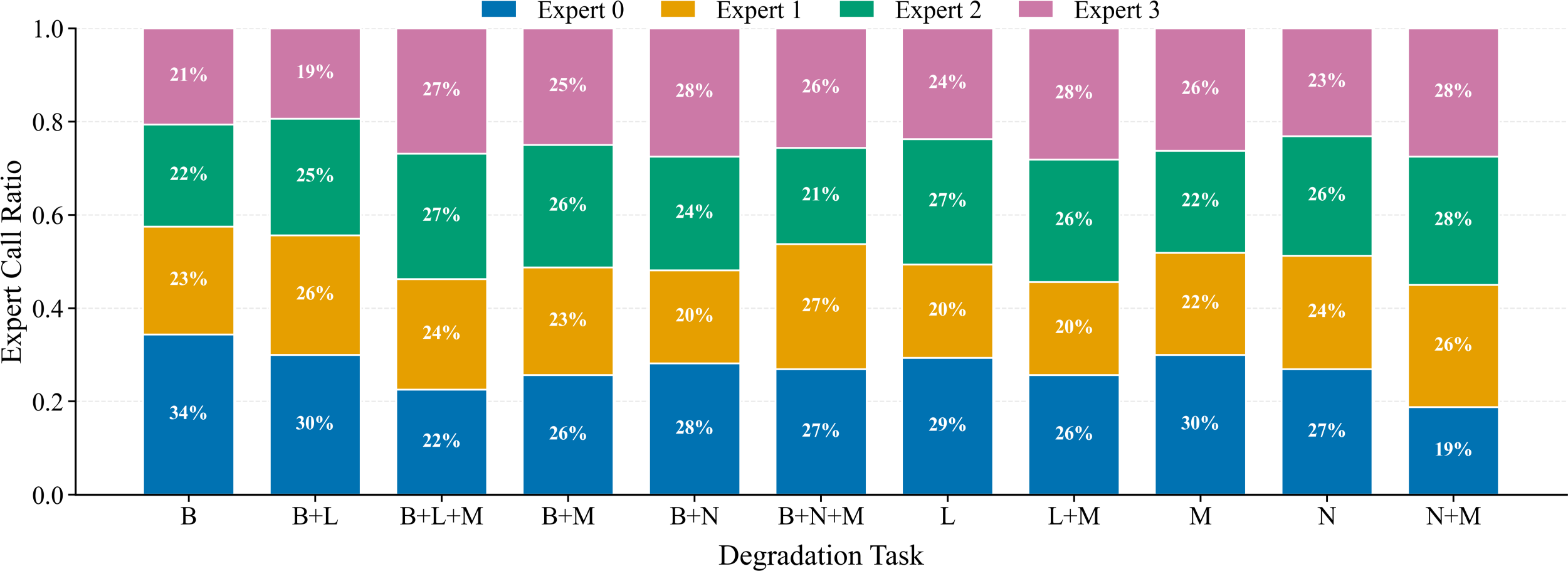}
	\end{center}
	\vspace{-1.5em}
	\caption{Activation patterns of each expert on PolarCDD.
	The probability indicates the proportion of
	images in a given task that are routed to
	the corresponding expert.}
	\label{fig:8}
\end{figure}

\subsection{MoE Analysis}
\label{A8}
Fig. \ref{fig:8} shows the activation patterns of
different MoE experts under various degradations.
All experts are effectively utilized while exhibiting
distinct specialization patterns.

Tabs. \ref{tab:10} and \ref{tab:11} report the effect of MoE under
the AIO3 setting and on PolarCDD. With MoE, our method achieves
consistent improvements across all metrics on AIO3.
Since the pretrained restoration prior does not
include demosaicking capability, the MoE extension
helps adapt the model to this unseen degradation without
disrupting the original prior.
On PolarCDD, MoE brings only limited gains
in intensity metrics but improves
polarization-related metrics.
This suggests that, under composite degradations,
the additional MoE capacity makes the
intensity features more transferable to the polarization domain,
thereby indirectly enhancing polarization restoration.
\subsection{Limitations and Future Work}
\label{A9}
We propose an AIO polarization restoration
model that achieves strong performance
across diverse degradation tasks. However,
its gains over task-specific expert models
remain marginal. In addition, the polarization
branch introduces extra computational
overhead, making our method particularly
demanding when built upon Transformer-based
backbones. This increased computational cost
is a trade-off for achieving stronger restoration performance.

In addition, the restoration priors
are still constrained by the distribution
and scale of general restoration datasets,
which remain relatively limited compared with
large-scale vision corpora. Leveraging
generative priors from large pretrained
models therefore represents a promising direction.
We have explored transferring a pretrained
text-to-image diffusion model (i.e., Stable Diffusion (\cite{rombach2022high}))
to polarization restoration. However, training
tends to become unstable when the polarization
branch remains lightweight, while scaling the
polarization branch to a capacity comparable to
Stable Diffusion introduces substantial deployment
costs. How to effectively incorporate large-scale
generative priors into polarization restoration
therefore remains an important direction for future research.
\subsection{Additional Visual Results}
\label{A10}

\begin{figure}[!t]
    \centering
    \includegraphics[width=0.95\linewidth]{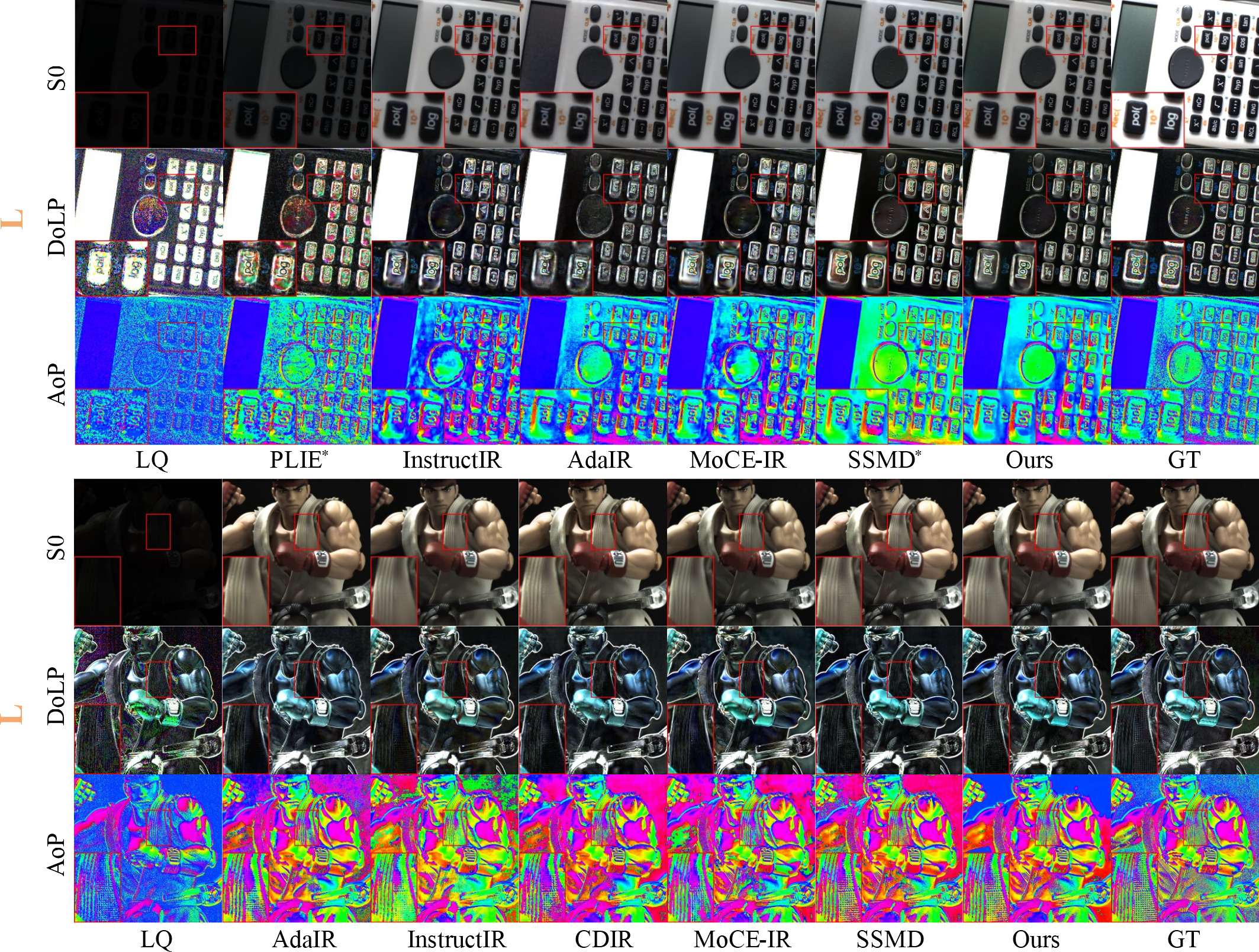}
    \vspace{-1.0em}
    \caption{Visual comparisons of
	different methods on the dataset LLCP (\cite{xu2022colorpolarnet}) and PolarCDD.
	L is low-light. The superscript * indicates task-specific weights.}
    \label{fig:9}
\end{figure}

\begin{figure}[!t]
    \centering
    \includegraphics[width=0.95\linewidth]{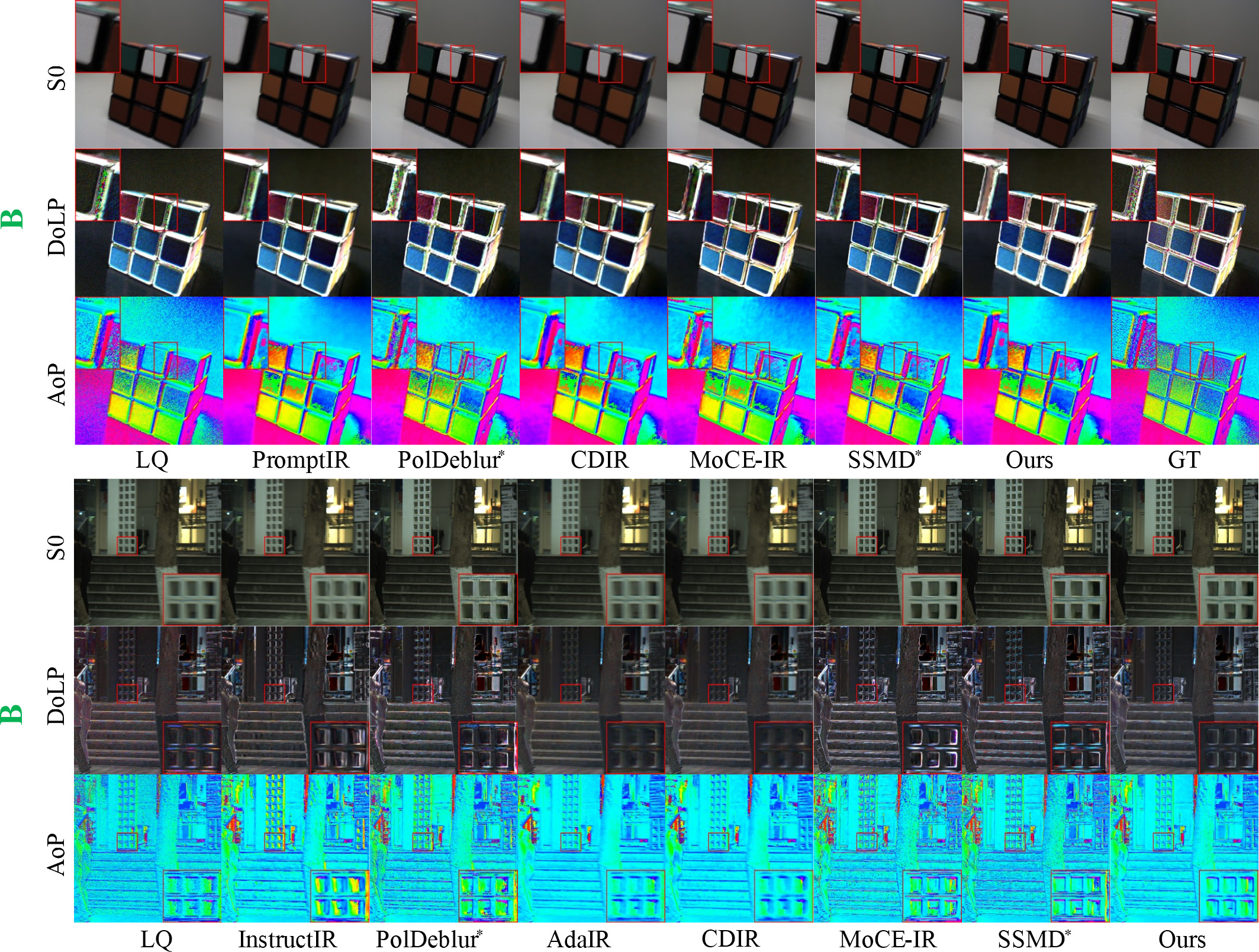}
    \vspace{-1.0em}
    \caption{Visual comparisons of
	different methods on the dataset PolDeblur (\cite{zhou2025learning}) and real-world dataset PV (\cite{li2026polarvsr}). 
	B is blur. The superscript * indicates task-specific weights.}
    \label{fig:10}
\end{figure}

\begin{figure}[!t]
    \centering
    \includegraphics[width=0.75\linewidth]{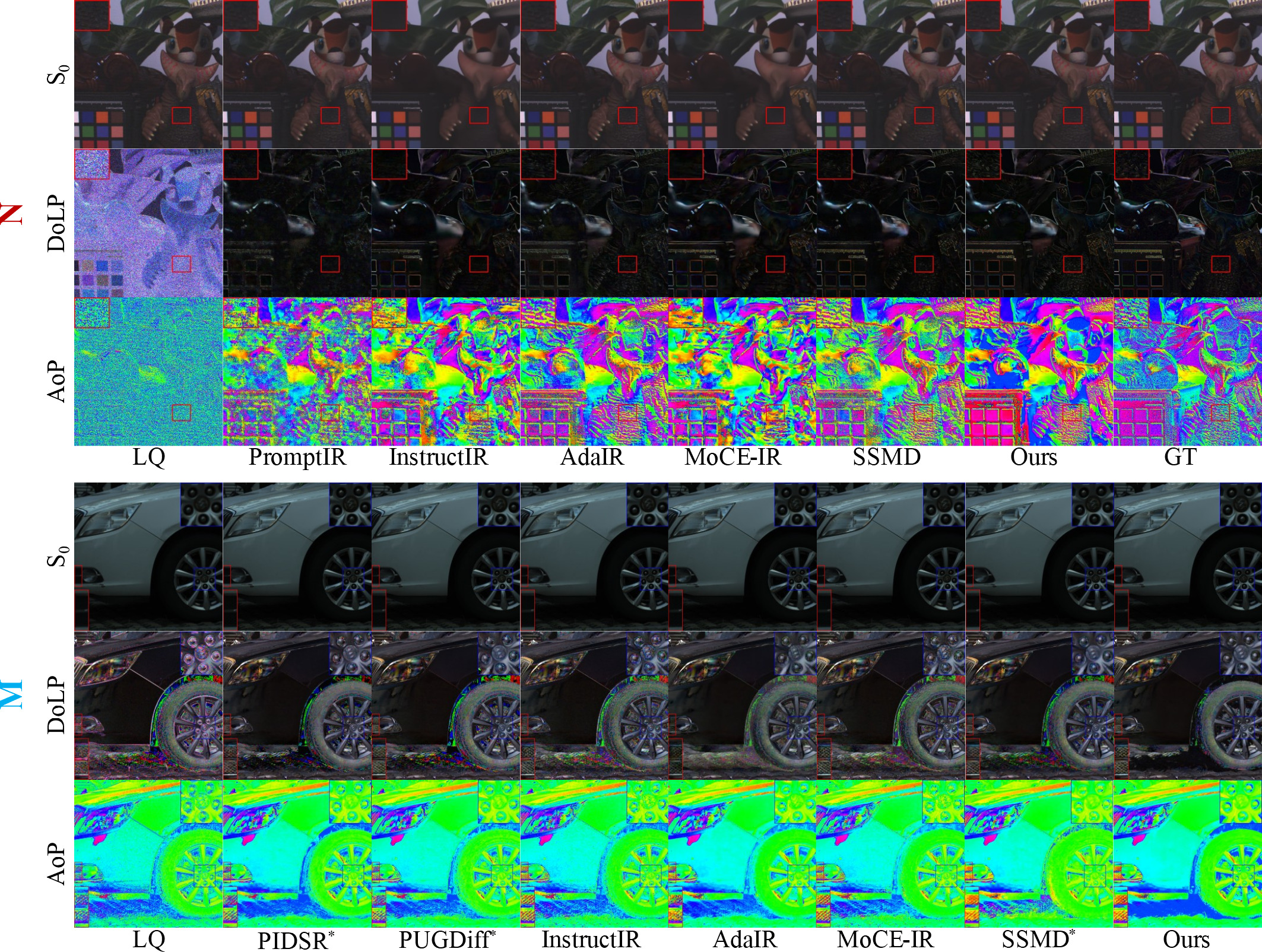}
    \vspace{-1.0em}
    \caption{Visual comparisons of
	different methods on the dataset PolarCDD and real-world dataset (\cite{luo2025cpifuse}).
	N is noise and M is mosaic. The superscript * indicates task-specific weights.}
    \label{fig:11}
\end{figure}

\begin{figure}[!t]
    \centering
    \includegraphics[width=0.75\linewidth]{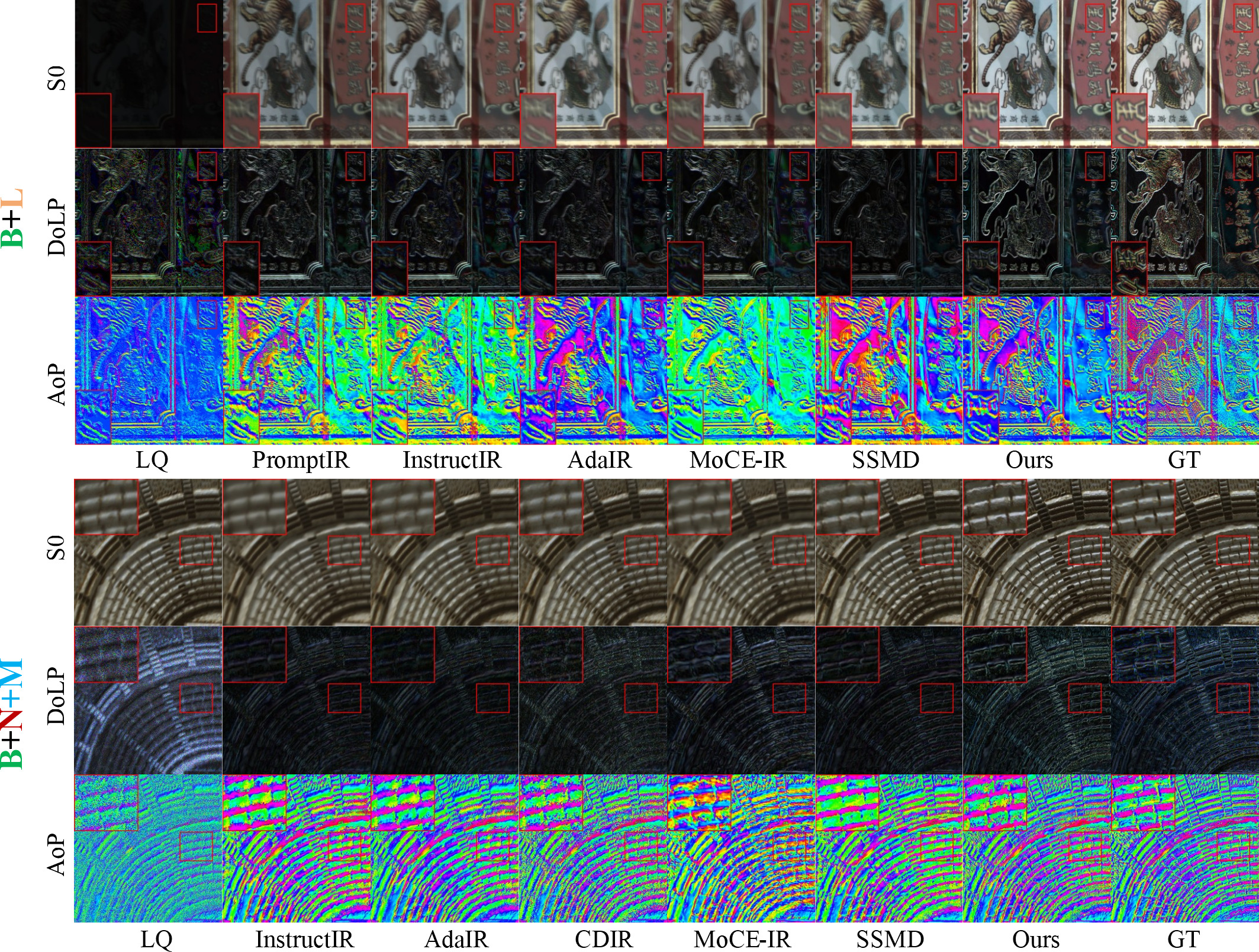}
    \vspace{-1.0em}
    \caption{Visual comparisons of
	different methods on the dataset PolarCDD.
	B is blur, L is low light, N is noise and M is mosaic.}
    \label{fig:12}
\end{figure}

Figs. \ref{fig:9}-\ref{fig:12} present additional qualitative
results. For deblurring (Fig. \ref{fig:10}), we
evaluate our method on the real-world PV
dataset (\cite{li2026polarvsr}), a vehicle-mounted polarization dataset
in which motion blur frequently occurs. Our method
removes motion blur more effectively than the
compared approaches. We also evaluate on
a real-world polarization mosaic dataset ((\cite{luo2025cpifuse}), Fig. \ref{fig:11}),
where our method produces the cleanest results and
achieves visual quality comparable to task-specific
expert models. Fig. \ref{fig:12} further presents results
under composite degradations, showing that our method
suppresses polarization-domain degradations more
effectively than other AIO restoration models.

\begin{figure}[!t]
    \centering
    \includegraphics[width=\linewidth]{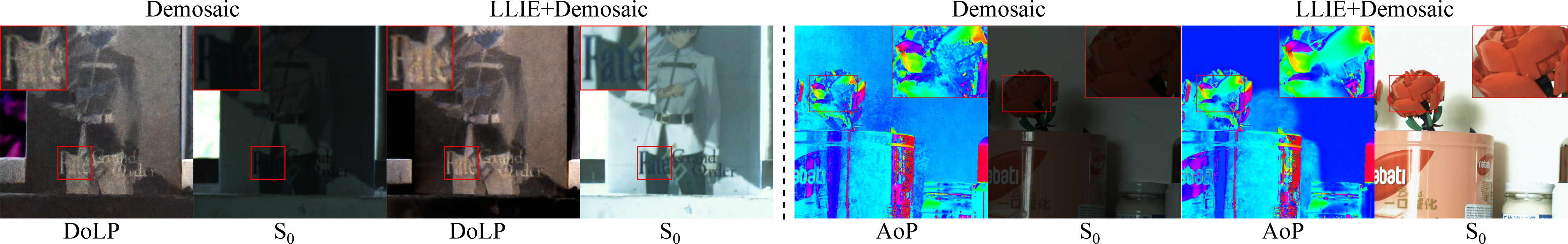}
    \vspace{-1.0em}
    \caption{Text-conditioned restoration behavior.}
    \label{fig:13}
\end{figure}

In addition, Fig. \ref{fig:13} demonstrates the
necessity of jointly addressing composite
degradations. Simultaneous low-light
enhancement and demosaicking better recover
image textures while reducing noise in the
polarization parameters.

\FloatBarrier
\end{document}